\PassOptionsToPackage{dvipsnames,table}{xcolor}

\documentclass[10pt,twocolumn,letterpaper]{article}

\usepackage{graphicx}     
\usepackage{subcaption}

\usepackage{array}
\usepackage{mathtools}
\usepackage{microtype}
\usepackage{bbm}

\usepackage{tcolorbox}
\tcbuselibrary{breakable}
\usepackage{dirtree}
\usepackage{booktabs} 
\usepackage{tabularx}
\usepackage{amssymb}
\usepackage{listings}
\usepackage{multirow}  
\usepackage[pagenumbers]{cvpr} 

\definecolor{eclipseBlue}{RGB}{42,0.0,255}
\definecolor{eclipseGreen}{RGB}{63,127,95}
\definecolor{eclipsePurple}{RGB}{127,0,85}
\definecolor{lbcolor}{rgb}{0.95,0.95,0.95} 

\lstdefinelanguage{json}{
    basicstyle=\ttfamily\footnotesize, 
    breaklines=true,                   
    breakatwhitespace=false,           
    showstringspaces=false,            
    commentstyle=\color{eclipseGreen}, 
    keywordstyle=\color{eclipsePurple}\bfseries, 
    stringstyle=\color{eclipseBlue},   
    literate=
     *{0}{{{\color{eclipseBlue}0}}}{1}
      {1}{{{\color{eclipseBlue}1}}}{1}
      {2}{{{\color{eclipseBlue}2}}}{1}
      {3}{{{\color{eclipseBlue}3}}}{1}
      {4}{{{\color{eclipseBlue}4}}}{1}
      {5}{{{\color{eclipseBlue}5}}}{1}
      {6}{{{\color{eclipseBlue}6}}}{1}
      {7}{{{\color{eclipseBlue}7}}}{1}
      {8}{{{\color{eclipseBlue}8}}}{1}
      {9}{{{\color{eclipseBlue}9}}}{1}
      {:}{{{\color{black}:}}}{1}       
      {,}{{{\color{black},}}}{1}       
      {\{}{{{\color{black}\{}}}{1}     
      {\}}{{{\color{black}\}}}}{1}
      {[}{{{\color{black}[}}}{1}       
      {]}{{{\color{black}]}}}{1},
}

\definecolor{cvprblue}{rgb}{0.21,0.49,0.74}
\definecolor{skyblue}{RGB}{30, 144, 255}
\usepackage[pagebackref,breaklinks,colorlinks,allcolors=cvprblue]{hyperref}

\usepackage{balance}

\def\paperID{815} 
\def\confName{CVPR}
\def\confYear{2026}
\definecolor{lightgrey}{gray}{0.9}
\definecolor{deepgrey}{gray}{0.75}
\title{PosterCopilot: Toward Layout Reasoning and Controllable Editing for Professional Graphic Design}

\author{
    Jiazhe Wei$^{1*}$, Ken Li$^{1*}$, Tianyu Lao$^{2}$, Haofan Wang$^{2}$, Liang Wang$^{1,3}$, Caifeng Shan$^{1}$, Chenyang Si$^{1 \dagger}$  \\[2mm] %
    $^{1}$PRLab, Nanjing University \quad $^{2}$LibLib.ai \quad  $^{3}$ Institute of Automation, Chinese Academy of Sciences \\ [2mm]
   \textbf{Project Page:} \href{https://postercopilot.github.io/}{\color{skyblue}\textbf{https://postercopilot.github.io/}}
}

\begin{document}

\twocolumn[{%
\renewcommand\twocolumn[1][]{#1}%
\maketitle

\vspace{-9mm} 
\includegraphics[width=1\linewidth]{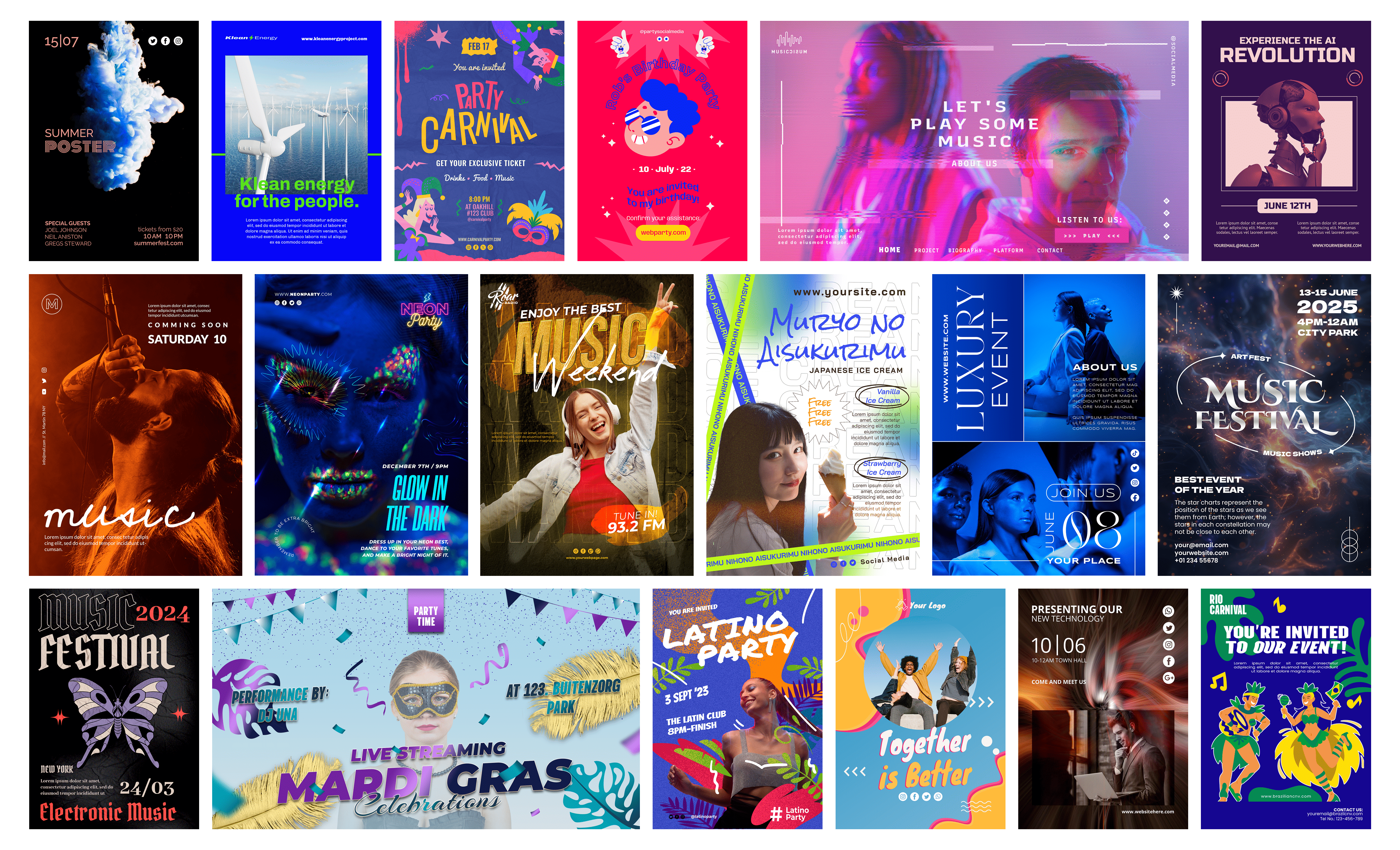}
\captionof{figure}{Generated results from our \textbf{PosterCopilot}. PosterCopilot exhibits exceptional graphic design capabilities by creating artworks with professional-grade layout, compelling visuals, and cohesive themes. \\[0pt]}

\label{fig:teaser}
}]
\vspace{-3mm}

{
    \renewcommand{\thefootnote}{\fnsymbol{footnote}}
    \footnotetext[1]{Equal Contribution}
    \footnotetext[2]{Corresponding author (chenyang.si@nju.edu.cn)} 
}

\begin{abstract}

Graphic design forms the cornerstone of modern visual communication, serving as a vital medium for promoting cultural and commercial events. Recent advances have explored automating this process using Large Multimodal Models (LMMs), yet existing methods often produce geometrically inaccurate layouts and lack the iterative, layer-specific editing required in professional workflows. To address these limitations, we present PosterCopilot, a framework that advances layout reasoning and controllable editing for professional graphic design. Specifically, we introduce a progressive three-stage training strategy that equips LMMs with geometric understanding and aesthetic reasoning for layout design, consisting of Perturbed Supervised Fine-Tuning, Reinforcement Learning for Visual-Reality Alignment, and Reinforcement Learning from Aesthetic Feedback. Furthermore, we develop a complete workflow that couples the trained LMM-based design model with generative models, enabling layer-controllable, iterative editing for precise element refinement while maintaining global visual consistency. Extensive experiments demonstrate that PosterCopilot achieves geometrically accurate and aesthetically superior layouts, offering unprecedented controllability for professional iterative design. 

\end{abstract}   

\section{Introduction}
\label{sec:intro}
\vspace{-1mm}

Graphic design serves as a fundamental medium for visual communication \cite{meggs2025meggs}, translating abstract ideas into clear and engaging visuals. It brings together images, text, and graphic elements in a deliberate way to create layouts that are both informative and visually appealing, bridging creativity with effective communication \cite{baldwin2006visual}. Recently, growing interest has emerged in automating the graphic design process through artificial intelligence. One major line of work explores diffusion-based generative models, which leverage their strong image synthesis capabilities to create visually rich posters \cite{li2023gligen,huang2024layerdiff,wang2024ms}. However, because these models generate all image regions simultaneously, they struggle to preserve the structural integrity, texture fidelity, and stylistic consistency of user-provided assets, making local refinements prone to distortion \cite{hou2024high,malhi2025preserving}. Another line leverages Large Multimodal Models (LMMs) to reason over design elements and predict their spatial and layer-wise arrangements, determining each element’s position, scale, and ordering within the composition \cite{wang2024divide,patnaik2025aesthetiq,hurst2024gpt}. These methods preserve the authenticity of visual assets and introduce interpretability and controllability into the design process, representing a promising step toward layout-centric and automation-oriented graphic design.

\begin{figure}[ht]
    \centering
    \includegraphics[width=\columnwidth]{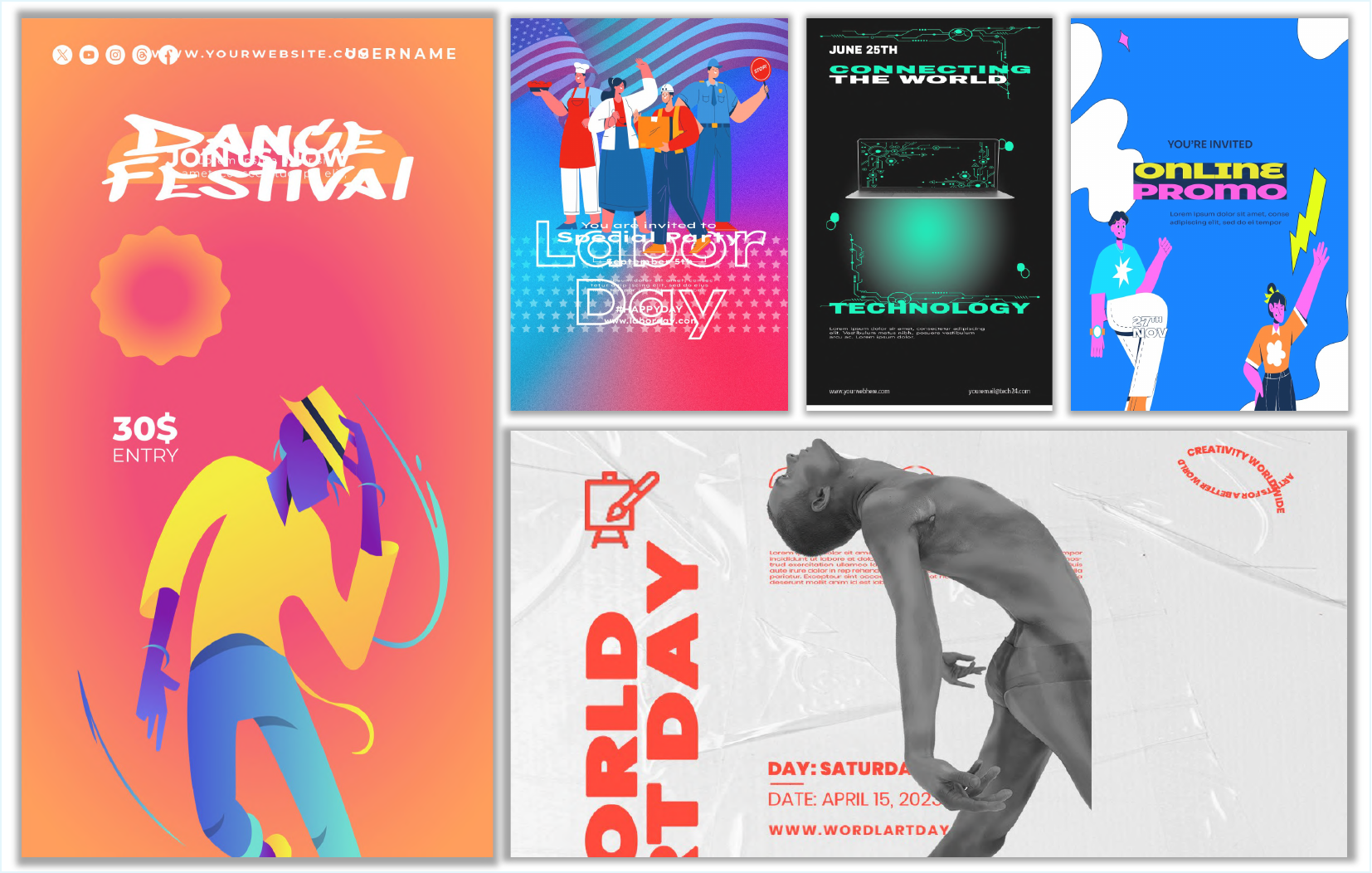}
    \captionsetup{aboveskip=0pt,belowskip=-18pt}
    \caption{Some failure cases created by existing design models in real-world, multi-asset scenarios, producing severe misalignments and visual discord.}
    \label{fig:intro_bad_case} 
\end{figure}

Despite these advances, current LMM-based methods still exhibit notable limitations when applied to professional design workflows: 1) when handling complex and numerous assets, existing methods often produce inaccurate and unaesthetic layouts \cite{liu2024visualagentbench,wang2025bridging} as shown in Fig.~\ref{fig:intro_bad_case}. We identify that existing methods rely on supervised fine-tuning (SFT) over discrete textual tokens to represent continuous spatial coordinates, creating a mismatch between the model’s symbolic representation and the true Euclidean geometry of layout design \cite{li2025llms,jiang2025detect}. This mismatch leads to misalignment, distortion, and suboptimal compositions. Moreover, these models lack visual feedback during training, which limits their ability to perceive and reason about aesthetic layouts \cite{patnaik2025aesthetiq,qu2025silmm}. 2) More critically, current LMM-based approaches merely generate initial drafts and lack interactive editing capabilities \cite{khan2025beyond,muehlhaus2024interaction}. However, professional designers refine the drafts through multiple rounds of precise, layer-specific adjustments \cite{janusz2025one,ocampo2024integrating,cha2025understanding}. Therefore, enabling iterative refinement is a crucial requirement for advancing AI-assisted graphic design toward practical applications \cite{karimi2020creative,oksanen2024bridging}.

To address these challenges, we propose \textbf{PosterCopilot}, which advances the field toward layout reasoning and controllable editing for professional graphic design. Specifically, to mitigate the inaccurate and unaesthetic layouts resulting from token-based coordinate representations, we propose Perturbed Supervised Fine-Tuning (PSFT), which reformulates coordinate regression into a distribution-based learning paradigm by introducing controlled perturbations to ground-truth coordinates. Compared to point-wise regression, learning a coordinate distribution allows the model to reason over continuous spatial relationships rather than memorizing discrete positions \cite{wu2025llm,li2024spatial}, leading to more coherent and aesthetically balanced layouts. To further address the lack of visual feedback and aesthetic understanding, we introduce a two-stage reinforcement learning (RL) strategy. In the first stage, Reinforcement Learning for Visual-Reality Alignment (RL-VRA) introduces verifiable geometric reward signals to explicitly correct residual spatial inaccuracies after SFT. In the second stage, Reinforcement Learning from Aesthetic Feedback (RLAF) employs a learned aesthetic reward model to encourage the model to generate aesthetically coherent and diverse compositions that extend beyond the ground truth.

Finally, to enable iterative and controllable refinement beyond initial generation, we develop a complete workflow that couples the trained LMM-based design model with the generative models, seamlessly integrating asset creation with precision editing. This workflow supports layer-specific, iterative editing, allowing precise modification of individual elements while maintaining global visual consistency. It empowers designers with multi-round, high-fidelity editing capabilities, enabling flexible adjustments to specific layers without altering surrounding content. Experimental results indicate that the design model trained via our three-stage method produces layouts that are both accurate and visually appealing, even rivaling or surpassing the Nano-Banana. More significantly, PosterCopilot's integration of a generative agent provides precise layer-wise editing. This transforms it into a powerful assistant, allowing designers to take a well-composed draft as a starting point and have it further optimized for enhanced aesthetics and practical application.

Our main contributions are summarized as follows:
\begin{itemize}
   \item We propose PosterCopilot, the first framework to decouple complex poster design into layout reasoning and multi-round lossless editing, demonstrating exceptional capabilities in both aspects.
    \item We introduce a progressive alignment training paradigm (PSFT, RL-VRA, RLAF) that enables LMMs to reason over continuous spatial relationships while instilling design principles and human aesthetics.
   \item We design a generative agent that supports iterative, controllable refinement beyond the initial generation, empowering PosterCopilot to serve as a powerful assistant for real-world editing scenarios.
    \item We contribute a large-scale, high-quality multi-layer poster dataset with rational granularity, along with its construction pipeline, addressing critical gaps in data scarcity and layer segmentation to benefit future research and applications.
\end{itemize}

\section{Related Work}
\label{sec:related_work}
\vspace{-1mm}

Multi-layer Graphic Layout Planning prioritizes real-world practicality by first inferring layouts, then assembling layers for optimal flexibility. LMM-assisted approaches (LayoutPrompter~\cite{lin2023layoutprompter}, LayoutNUWA~\cite{tang2023layoutnuwa}, PosterLLaVA~\cite{yang2024posterllava}) employed in-context learning, while others specialized in asset integration (Graphist~\cite{cheng2025graphic}), typography (POSTA~\cite{chen2025posta}), or external generation (CreatiPoster~\cite{zhang2025creatiposter}, COLE~\cite{jia2023cole}). Crucially, these methods mimic static datasets rather than learning from aesthetic outcomes. Our approach transcends limitations by internalizing layout principles and visual aesthetics through direct generative feedback. \textcolor[HTML]{F52731}{\textbf{\textit{More discussion is in supplementary material.}}}

\section{Methodology}
\label{sec:Method_1}
\vspace{-1mm}

\begin{figure}[t!]
    \centering 
    
    \begin{subfigure}[b]{0.95\linewidth} 
        \centering
        \includegraphics[width=\linewidth]{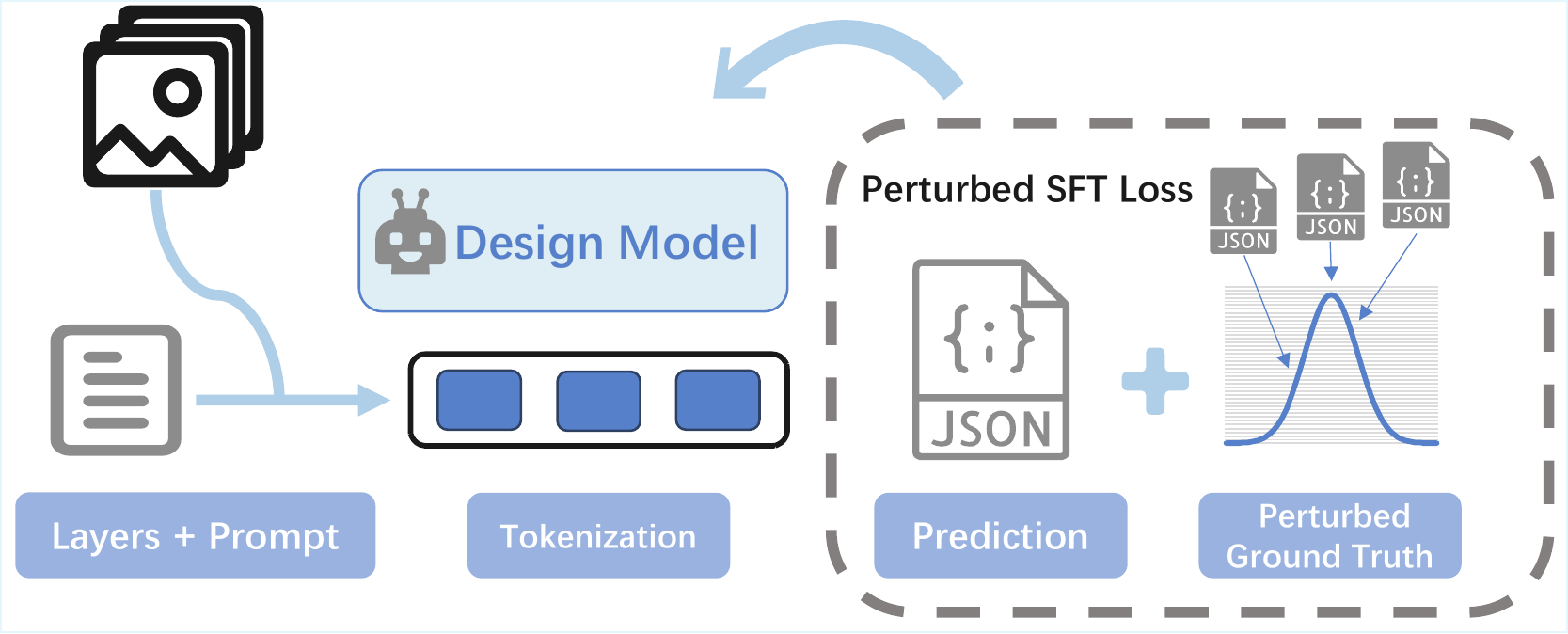} 
        \caption{Perturbed Supervised Fine-Tuning (PSFT)}
        \label{fig:train_psft}
    \end{subfigure}
    
    \vspace{\baselineskip}

    \begin{subfigure}[b]{0.95\linewidth}
        \centering
        \includegraphics[width=\linewidth]{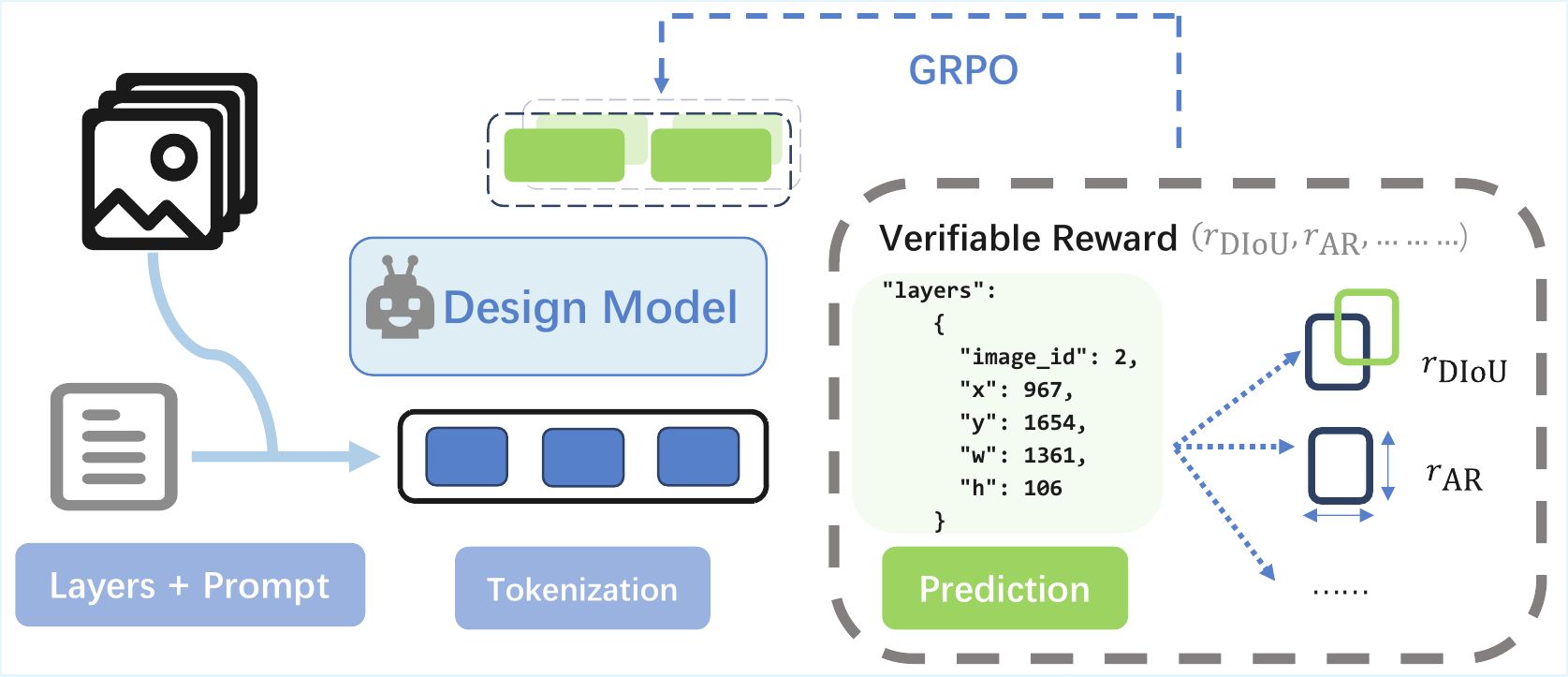}
        \caption{Reinforcement Learning for Visual-Reality Alignment (RL-VRA)}
        \label{fig:train_rlvra}
    \end{subfigure}

    \vspace{\baselineskip} 

    \begin{subfigure}[b]{0.95\linewidth}
        \centering
        \includegraphics[width=\linewidth]{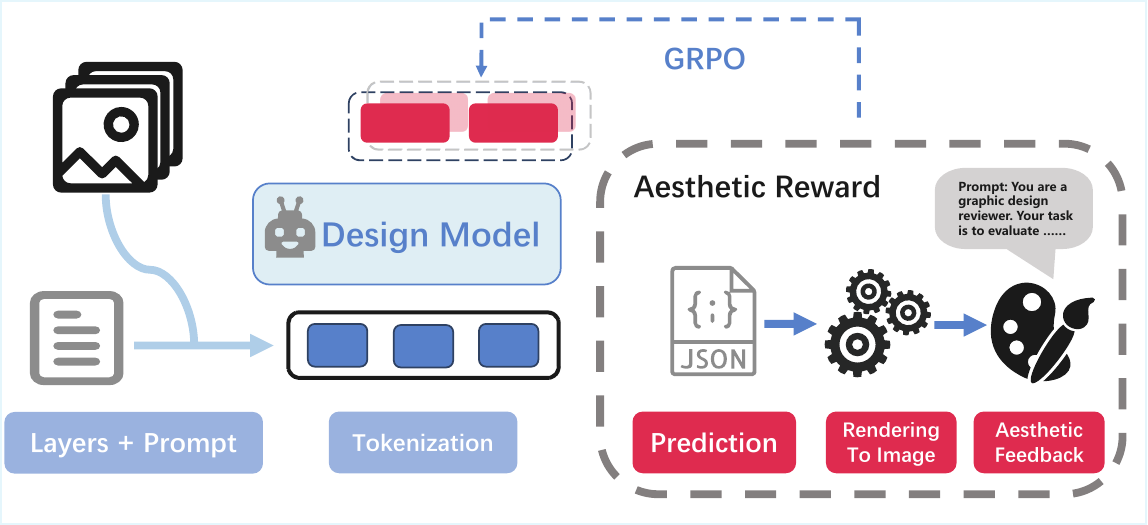}
        \caption{Reinforcement Learning from Aesthetic Feedback (RLAF)}
        \label{fig:train_rlaf}
    \end{subfigure}

    \captionsetup{aboveskip=2pt,belowskip=-14pt}
    \caption{\textbf{Overview of the training paradigm of PosterCopilot.} Rather than formulating the training process as a simple regression task, we endow PosterCopilot with outstanding layout capabilities and human-like aesthetics through a three-stage training paradigm.}
    \label{fig:training_paradigm}
\end{figure}

In this section, we will first detail the training paradigm for the design model, and subsequently present the complete PosterCopilot pipeline. Our three-stage design model training paradigm is illustrated in Fig.~\ref{fig:training_paradigm}. 

\subsection{Task Formulation}
\vspace{-1mm}

Our objective is to automatically arrange user-provided elements $E=\{e_1, \dots, e_N\}$ of types $\mathcal{T}=\{\text{image}, \text{text}, \text{shape}\}$ on a canvas, achieving aesthetic coherence while preserving asset fidelity. Text elements are rasterized into image layers for unified processing. The input elements and canvas dimensions $(H_c, W_c)$ are encoded into a multimodal prompt $P^{E}_{H_c, W_c}$, which our design model $\mathcal{M}$ processes to generate the final layout:

\begin{equation}\label{eq:layout_gen}
\mathcal{M}(P^{E}_{H_c, W_c}) \to \boldsymbol{G}
\end{equation}
where $\boldsymbol{G} = \{(\boldsymbol{b}_i, l_i)\}_{i=1}^N$ specifies each element's bounding box $\boldsymbol{b}_i$ and layer order $l_i$.

\subsection{Perturbed Supervised Fine-Tuning}
\vspace{-1mm}

\begin{figure}[ht]
    \centering
    \includegraphics[width=1\columnwidth]{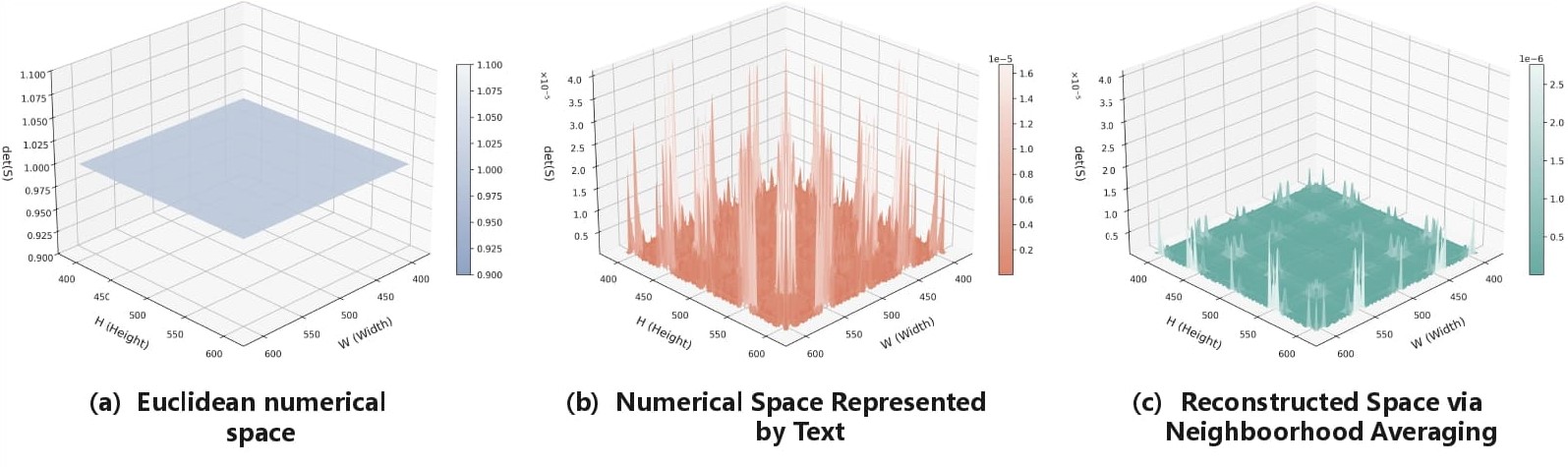} 
    \captionsetup{aboveskip=0pt,belowskip=-7pt}
    \caption{Geometric instability of text-based coordinate representations. (a) \textbf{Euclidean Space:} The ideal baseline, showing perfect, uniform geometry ($\det(S) \equiv 1$). (b) \textbf{Text-Based Space:} Suffers from signal collapse (near-zero $\det(S)$) and geometric noise, creating a chaotic landscape unstable for optimization. (c) \textbf{Reconstructed Space via Neighborhood Averaging:} This method suppresses noise, recovering a smooth, uniform geometry that is far more stable than (b).}
    \label{fig:motivation_sft} 
\end{figure}

We posit that the standard LMM practice of quantizing continuous coordinates into discrete text tokens fundamentally warps the optimization space's geometry \cite{davies2025language, karthick2025LLMReg, inoue2023layoutdm}, hindering precise localization. To validate this, we visualize the local geometric uniformity using $det(S)$, the determinant of the Structure Tensor $S$ \cite{harris1988combined,bigun2002multidimensional}. As shown in Fig.~\ref{fig:motivation_sft}, the ideal Euclidean space (a) has $\det(S) \equiv 1$, whereas the text-represented numerical space (b) is geometrically broken. Critically, (c) confirms that neighborhood averaging---our core insight---effectively repairs this distortion and recovers a stable optimization signal.

Based on this finding, we propose \textbf{Perturbed Supervised Fine-Tuning (PSFT)}. Instead of point-wise regression on ground-truth layout $\boldsymbol{G}_{\text{gt}} = \{(\boldsymbol{b}_i, l_i)\}_{i=1}^N$, we sample $n$ perturbed variants $\boldsymbol{G}_{\text{pert}}^{(i)}$ by injecting Gaussian noise specifically on the bounding box values $b_i$:
\begin{equation}
\label{eq:noise}
\boldsymbol{G}_{\text{pert}}^{(i)} \sim \mathcal{N}(\boldsymbol{G}_{\text{gt}}, \sigma^2 \mathbf{I}), \quad i = 1, 2,\dots, n
\end{equation}
where $\sigma$ is a small standard deviation. Our training objective, $\mathcal{L}_{\text{PSFT}}$, combines the standard cross-entropy loss on the original layout with an averaged loss over $n$ perturbations:
\begin{equation}
\label{eq:sft-loss}
\begin{split}
\mathcal{L}_{\text{PSFT}} = {} & L_{CE}(\hat{\boldsymbol{G}}, \boldsymbol{G}_{\text{gt}}) \\
 & + \lambda_{\text{Perturbed}} \cdot \frac{1}{n} \sum_{i=1}^n L_{CE}(\hat{\boldsymbol{G}}, \boldsymbol{G}_{\text{pert}}^{(i)})
\end{split}
\end{equation}
where $\hat{\boldsymbol{G}}$ is the model's prediction. This formulation compels the model to learn a continuous spatial distribution centered on the ground truth, rather than memorizing discrete token positions, thereby mitigating the limitations of text-based regression.

\subsection{Reinforcement Learning for Visual-Reality Alignment}
\label{sec:rlvra}
\vspace{-1mm}

\begin{figure}[t]
    \centering
    \includegraphics[width=\columnwidth]{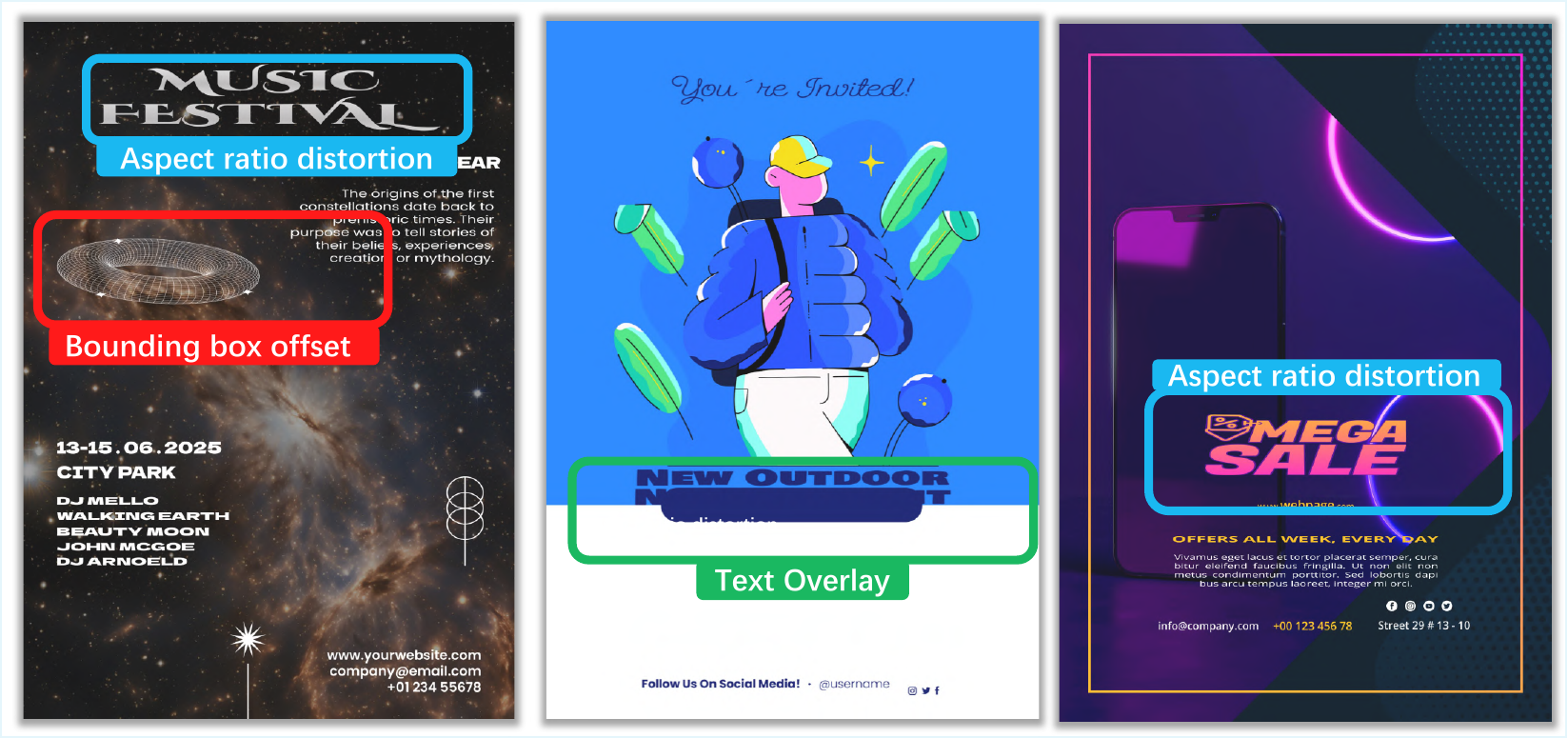}
    \captionsetup{aboveskip=0pt,belowskip=-14pt}
    \caption{Our motivation for visual-reality alignment and aesthetic feedback stems from the observation that design models frequently produce works that violate fundamental graphic design principles, as well as exhibit serious aesthetic flaws. We use \textcolor[HTML]{FF0000}{red}, \textcolor[HTML]{00B050}{green}, and \textcolor[HTML]{00B0F0}{blue} boxes to mark the error areas in the figure. }
    \label{fig:motivation_rlvr}
\end{figure}

\begin{figure*}[t!]
    \centering 
    
    \includegraphics[width=\textwidth]{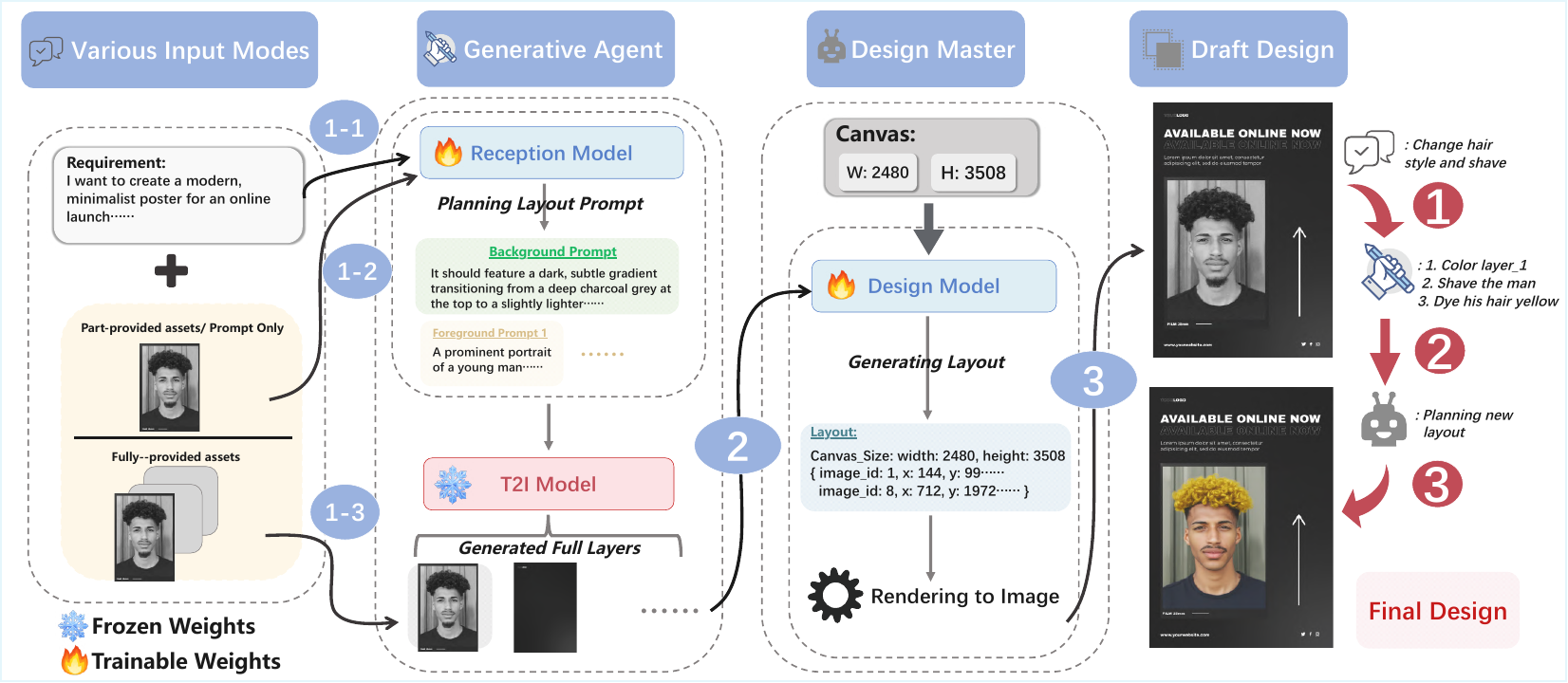}
    \caption{\textbf{Overview of PosterCopilot's Inference and Editing Pipeline.} The \textit{\textcolor[HTML]{8faadc}{standard inference}} and \textit{\textcolor[HTML]{ba3844}{multi-round editing}} pipelines are marked by \textcolor[HTML]{8faadc}{blue} and \textcolor[HTML]{ba3844}{red} numbers, respectively. Before layout design, PosterCopilot can supplement new assets when design materials are insufficient. \textbf{Generative Agent}  first processes user requirements, undergoes professional planning, and delivers complete assets. \textbf{Design Master}  then generates optimal compositions based on the assets and requirements, ultimately rendering the \textbf{Draft design.} The draft design will be revised into the \textbf{final design} after multiple rounds of editing by the collaboration of both generative agent and design master.}
    \label{fig:inference_pipeline}
\end{figure*}
While PSFT offers a robust spatial prior, its dependence on supervised learning without visual feedback results in geometric flaws, such as bounding box drift and aspect ratio distortion. Critically, these rendering-stage errors, evident in Fig.~\ref{fig:motivation_rlvr}, cannot be captured easily within the SFT paradigm itself. To bridge this \textbf{visual-reality gap} and align model outputs with graphic design principles, we introduce the \textbf{Reinforcement Learning for Visual-Reality Alignment (RL-VRA)} phase.

We frame RL-VRA as an online policy optimization task under a single-step Markov Decision Process (MDP). The state $s$ corresponds to the input prompt $P^{E}_{H_c, W_c}$, while the action $a$ represents the layout generation $\boldsymbol{G} = \{(\boldsymbol{b}_i, l_i)\}_{i=1}^N$. Our objective is to refine the pre-trained SFT policy $\pi_{\text{ref}}(\boldsymbol{G} \mid s)$ into an enhanced policy $\pi_\theta(\boldsymbol{G} \mid s)$ by maximizing the expected return under a geometry-aware reward signal:

\begin{equation}
\label{eq:optimization}
\begin{split}
    J_{VRA}(\theta) &= \mathbb{E}_{\boldsymbol{G} \sim \pi_\theta( \cdot \mid s)} \left[ r(\boldsymbol{G}) \right] \\
    & - \beta D_{KL}(\pi_\theta( \cdot \mid s) \,||\, \pi_{\text{ref}}(\cdot \mid s))
\end{split}
\end{equation}

where $J_{VRA}(\theta)$ balances reward maximization against policy conservatism, with $\pi_{\text{ref}}$ serving as the frozen reference policy, $\beta$ controlling the KL regularization strength \cite{schulman2017ppo}, and $r(\boldsymbol{G})$ providing dense \textbf{verifiable geometric visual feedback}. To ensure stable policy updates for high-dimensional discrete action spaces, we employ \textbf{Group Relative Policy Optimization (GRPO)} \cite{shao2024deepseekmath}, which operates without explicit value function estimation. For each group of $K$ policy rollouts, we compute:
\begin{equation}
\begin{split}
A_i &= r(\boldsymbol{G}_i) - \frac{1}{K} \sum_{j=1}^K r(\boldsymbol{G}_j) \\
r_i(\theta) &= \frac{\pi_\theta(\boldsymbol{G}_i \mid s)}{\pi_{\theta_{\text{old}}}(\boldsymbol{G}_i \mid s)}
\end{split}
\end{equation}

where $A_i$ represents the advantage of action $\boldsymbol{G}_i$ relative to the group, and $r_i(\theta)$ is the probability ratio between the new and old policies. Our reward function $r(\boldsymbol{G}) = r_{\text{Spatial}} + r_{\text{Element}}$ + $r_{\text{format}}$ provides multi-scale geometric supervision, decomposing layout quality into spatial coherence and element-level fidelity components.

The spatial reward $r_{\text{Spatial}}$ addresses layout misalignment through Distance Intersection over Union (DIoU) \cite{zheng2020diou}:
\begin{equation}
\label{diou_reward}
r_{\text{Spatial}} = r_{\text{DIoU}} = \sum_i \left( \text{IoU}(\boldsymbol{b}_i, \boldsymbol{b}_i^{\text{gt}}) - \frac{\boldsymbol{\rho}^2(\boldsymbol{b}_i, \boldsymbol{b}_i^{\text{gt}})}{\boldsymbol{c}^2} \right)
\end{equation}
where $\boldsymbol{\rho}$ denotes the center distance, $\boldsymbol{c}$ represents the diagonal of the minimal enclosing box, and $\boldsymbol{b}_i^{\text{gt}}$ is the ground-truth box from $\boldsymbol{G}_{\text{gt}}$.

The element-level reward $r_{\text{Element}} = r_{\text{AR}} + r_{\text{size}}$ penalizes geometric distortions that compromise visual integrity. The aspect ratio reward:
\begin{equation}
    \label{eq:ar}
    r_{\text{AR}} = -\sum_i \left| \log \left( \frac{w_i / h_i}{w_i^{\text{gt}} / h_i^{\text{gt}}} \right) \right|
\end{equation}
preserves element proportions, while the size reward:
\begin{equation}
\label{eq:size}
\resizebox{0.90\linewidth}{!}{%
    $\displaystyle r_{\text{size}} = -\sum_i \left[ \text{smooth}_{\delta} \left( \frac{w_i - w_i^{\text{gt}}}{w_i^{\text{gt}}} \right) + \text{smooth}_{\delta} \left( \frac{h_i - h_i^{\text{gt}}}{h_i^{\text{gt}}} \right) \right]$%
}
\end{equation}
maintains original dimensions using the Huber loss \cite{girshick2015fast}:
\begin{equation}
    \label{smooth}
    \text{smooth}_{\delta}(d) = \begin{cases} 0.5 d^2  / \delta & |d| < \delta \\ |d| - 0.5 \delta & \text{otherwise} \end{cases}
\end{equation}
where $\delta$ controls the transition between quadratic and linear regimes, preventing reward domination during extreme size distortions.

We further incorporate $r_{\text{format}}$ to enforce JSON-structured outputs. The complete reward formulation:
\begin{equation}
\label{eq:reward}
r(\boldsymbol{G}) = \underbrace{r_{\text{DIoU}}}_{\text{Spatial Coherence}} + \underbrace{\lambda_{\text{size}} r_{\text{size}} + \lambda_{\text{AR}} r_{\text{AR}}}_{\text{Element Fidelity}} + r_{\text{format}}
\end{equation}
where $\lambda_{\text{size}}, \lambda_{\text{AR}} > 0$ balance reward components. This geometrically-grounded reward structure injects explicit visual-reality constraints directly into the policy gradient updates, enabling the model to learn corrective behaviors that transcend the limitations of previous methods that lack visual feedback during training.

\subsection{Reinforcement Learning from Aesthetic Feedback}
\vspace{-1mm}

While prior stages enforce graphic design rules based on a single ground-truth, this is just one of many aesthetically valid solutions. To align with broader human aesthetic preferences, we introduce the \textbf{Reinforcement Learning from Aesthetic Feedback (RLAF)} stage. This stage explores a wider design space using a new subjective reward, $r_{\text{aes}}(\boldsymbol{G})$, provided by a pre-trained LMM (acting as an aesthetic judge) that evaluates the final \emph{rendered image}. This aesthetic score is combined with our format reward $r_{\text{format}}$:
\begin{equation}
\label{eq:rlaf_reward}
r_{\text{RLAF}}(\boldsymbol{G}) =r_{\text{format}} + \lambda_{\text{aes}} r_{\text{aes}}(\boldsymbol{G})
\end{equation}
where $\lambda_{\text{aes}} > 0$. This stage encourages the model to discover novel, high-appeal layouts that may surpass the ground-truth.

\subsection{Unleashing the Creative Flow: Generative Asset Synthesis and Iterative Refinement}
\label{sec:postercopilot_generative_agent}
\vspace{-1mm}

With our design model, we now unleash its creative potential by integrating a generative agent that completes the PosterCopilot framework. This integration transforms the model from a pure layout planner into a comprehensive design partner, capable of both asset synthesis and iterative editing. As shown in Fig.~\ref{fig:inference_pipeline}, this agent first addresses the issue of missing assets: when provided with only partial assets, it can adaptively generate new, style-consistent elements to complete the layout. Specifically, we utilize a trained LMM called the reception model to generate textual descriptions for each missing layer, which are then combined with existing assets as style reference images to be fed together into a text-to-image (T2I) model to generate the corresponding assets. More importantly, the generative agent supports fine-grained, multi-round editing required in professional workflows by accepting user instructions to perform targeted modifications on corresponding layers. This enables designers to perform stable, iterative cycles between 'precise single-layer asset editing' and 'global layout re-arrangement', while effectively mitigating common challenges in traditional editing methods, such as asset distortion and uncontrollable edit scopes.
\section{Application}
\vspace{-1mm}

Harnessing its powerful reasoning capability and fine-grained layer-wise architecture, PosterCopilot unlocks diverse applications in professional design scenarios.

\vspace{-1mm}

\begin{figure}
    \centering 
    
    \begin{subfigure}[b]{0.95\linewidth}
        \centering
        \includegraphics[width=\linewidth]{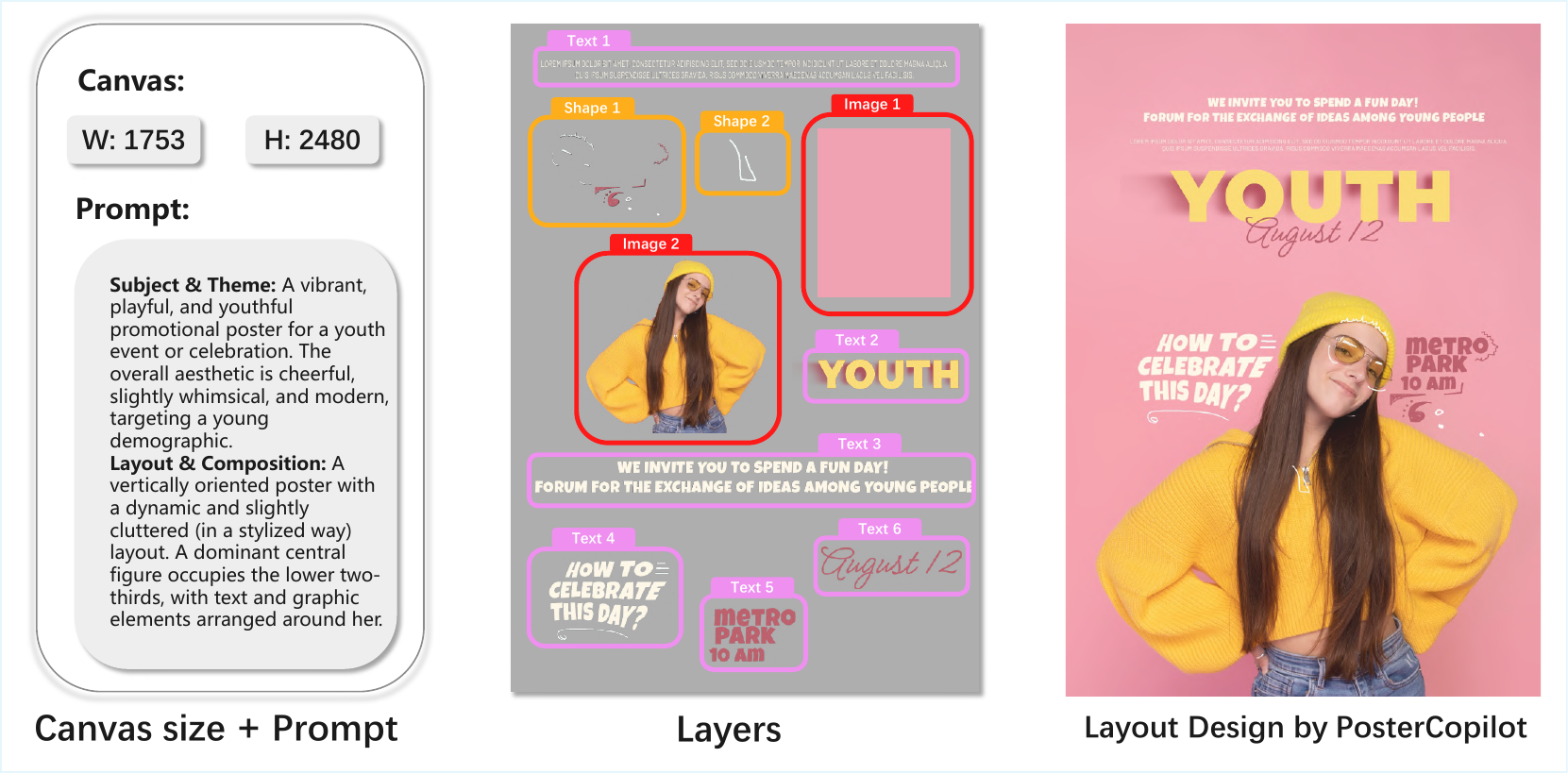}
        \label{fig:recon_2}
    \end{subfigure}
    \captionsetup{aboveskip=0pt, belowskip=-14pt}
    \caption{Poster generated from fully-provided assets by PosterCopilot.}
    \label{fig:recon_demo}
\end{figure}

\label{sec:application}
\subsection{Poster Generation from Fully-provided Assets}
\vspace{-1mm}

As shown in Fig.~\ref{fig:recon_demo}, PosterCopilot excels at arranging a complete set of user-provided assets into an aesthetically pleasing, professional-grade design, while guaranteeing every asset is faithfully preserved without alteration.

\vspace{-1mm}
\subsection{Poster Generation from insufficient Assets}
\vspace{-1mm}

\begin{figure*}[t!]
    \centering 
    \begin{subfigure}[b]{\textwidth} 
        \centering
        \includegraphics[width=\textwidth]{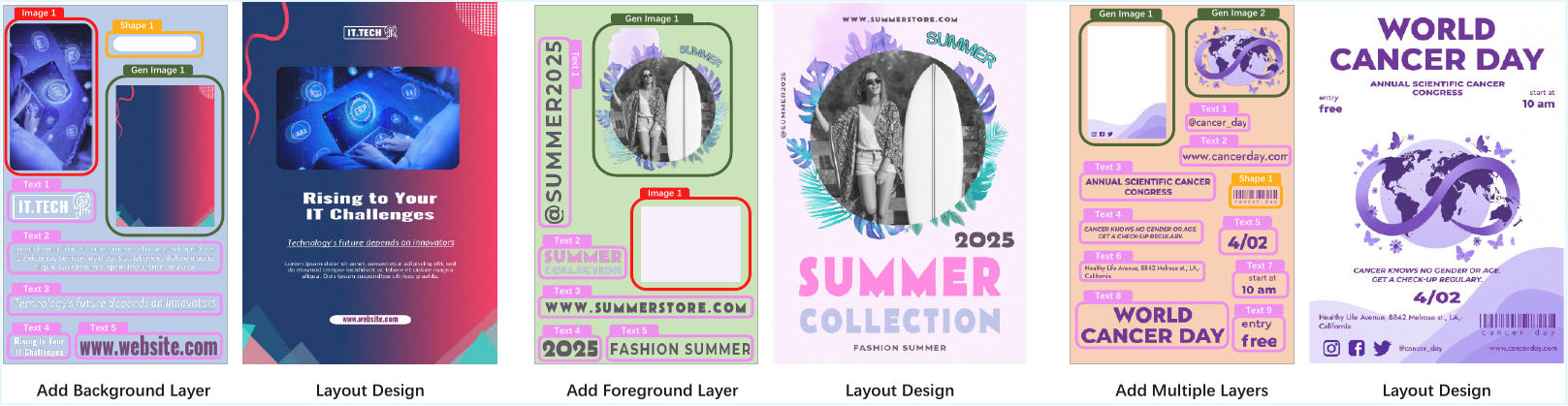}
    \end{subfigure}
    \captionsetup{aboveskip=4pt, belowskip=-4pt}
    \caption{Posters generated from insufficient assets by our PosterCopilot.}
    \label{fig:assets_complete} 
\end{figure*}

PosterCopilot's generative agent handles incomplete assets by synthesizing missing layers, such as background or foreground layers, with stylistic consistency. This capability, as shown in Fig.~\ref{fig:assets_complete}, accelerates the initial design phase by enabling rapid drafts generation where synthesized elements blend harmoniously with user-provided assets.
\begin{figure*}[t!]
    \centering 
    \begin{subfigure}[b]{\textwidth} 
        \centering
        \includegraphics[width=\textwidth]{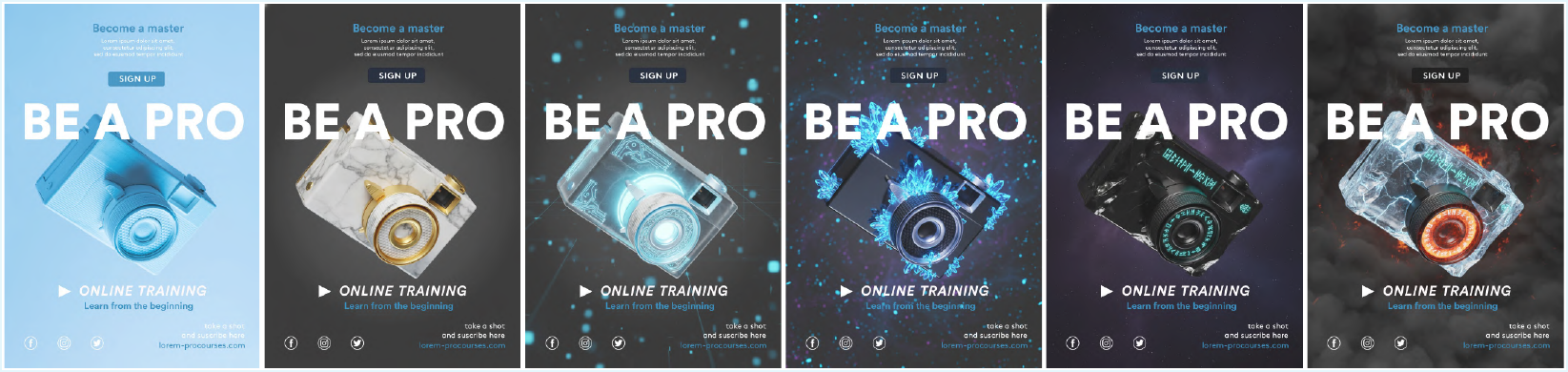}

        \label{fig:subim1} 
    \end{subfigure}
  
    \captionsetup{aboveskip=4pt, belowskip=-4pt}
    \caption{
        Multi-round refinement for a single layer by our PosterCopilot.
    }
    \label{fig:multi-round refine single object} 
\end{figure*}

\begin{figure*}[t!]
    \centering 
    \begin{subfigure}[b]{\textwidth} 
        \centering
        \includegraphics[width=\textwidth]{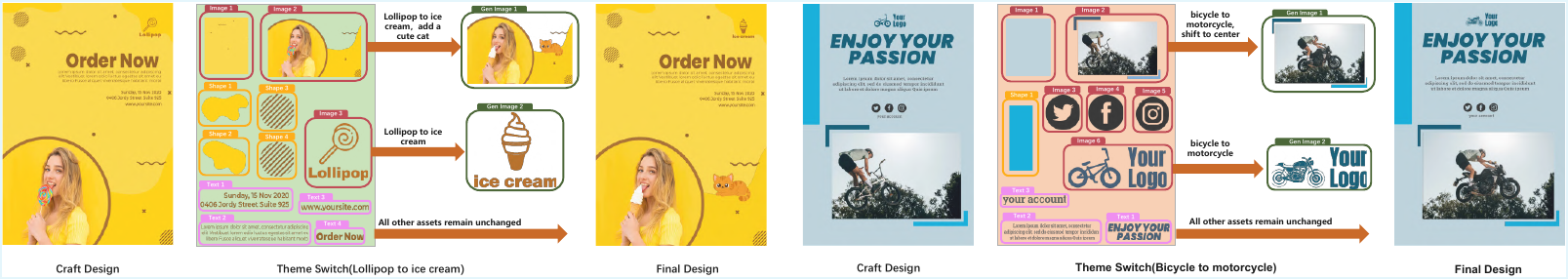}
    \end{subfigure}
    \captionsetup{aboveskip=4pt, belowskip=-7pt}
    \caption{
        Multi-round refinement for theme switch by our PosterCopilot.}
    \label{fig:multi-round refine theme switch}
\end{figure*}

\vspace{-1mm}
\subsection{Multi-round fine-grained Edit}
\vspace{-1mm}

PosterCopilot supports precise, multi-round editing of poster drafts. This functionality encompasses a diverse range of edit types, which we demonstrate in the following.
\subsubsection{Single Layer Edit}
\vspace{-1mm}

As shown in Fig.~\ref{fig:multi-round refine single object}, PosterCopilot supports multiple, varied edits on a single, fine-grained layer (e.g., modifying a camera's material or a character's pose). This high-fidelity process strictly confines the edit scope to the target layer, ensuring precise modification while preserving all other elements. This approach avoids the distortion common in diffusion-based methods that edit the entire poster.

\subsubsection{Theme Switch}
\vspace{-1mm}

Fig.~\ref{fig:multi-round refine theme switch} demonstrates the "Theme Switch" capability, enabling holistic theme migration through targeted, multi-round edits. For instance, users can swap "lollipop" elements for "ice cream," transforming the poster's theme (e.g., "lollipop sale" to "ice cream promotion") while perfectly preserving the original layout and decorative elements.

\subsubsection{Poster Reframe}
\vspace{-1mm}

\begin{figure*}[ht]
    \centering 
    \begin{subfigure}[b]{\textwidth} 
        \centering
        \includegraphics[width=\textwidth]{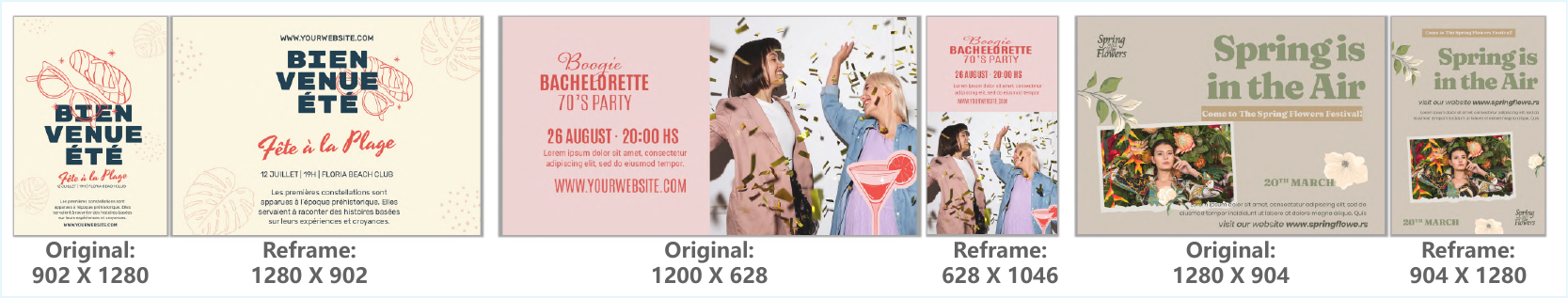}
    \end{subfigure}
    \caption{PosterCopilot intelligently reframes posters to new canvas sizes while maintaining layout harmony. All figures are scaled to a uniform height for presentation in this paper.}
    \label{fig:reframe}
\end{figure*}

Leveraging the design model's powerful reasoning capability, PosterCopilot can intelligently reframe and regenerate appropriate layouts simply by modifying the canvas size specification in the input requirements. Fig.~\ref{fig:reframe} presents examples of poster reframing by PosterCopilot.

\section{Experimental Details}
\label{sec:exp_details}
\vspace{-1mm}

\subsection{PosterCopilot Datasets}
\vspace{-1mm}

\begin{figure*}[t!]
    \centering 

        \includegraphics[width=\textwidth]{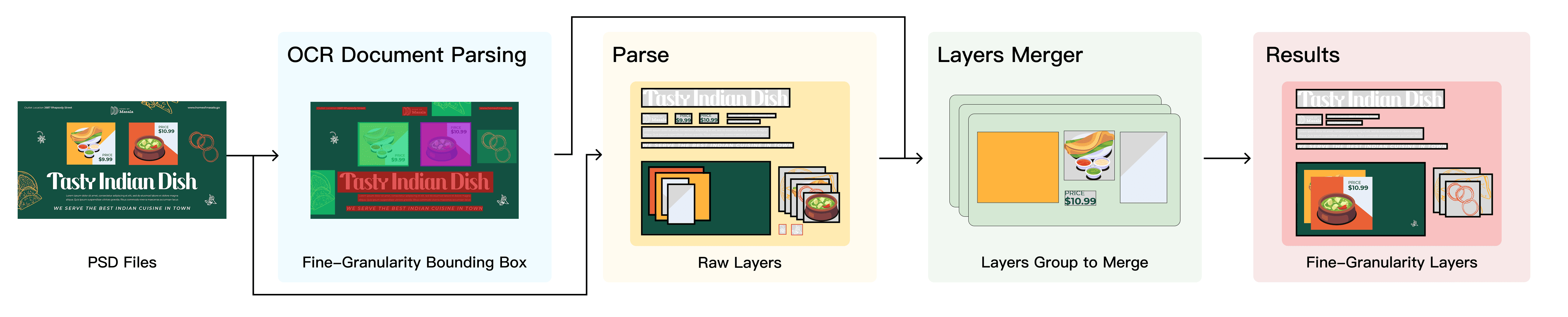}
       
    \caption{
        Dataset construction pipeline for our PosterCopilot. We merged numerous scattered layers with OCR-based fine-granularity bounding box rather than simply parsing the original PSD file.
    }
    \label{fig:data_cleaning_pipeline} 
\end{figure*}

A long-standing challenge in constructing high-quality, multi-layer poster datasets is over-segmentation, where a single visual element is fragmented across multiple independent layers \cite{zou2025fragment} (e.g., a shoe decomposed into separate layers for its laces, sole, and body). To solve this, we developed a novel construction pipeline. As illustrated in Fig. \ref{fig:data_cleaning_pipeline}, we employ OCR-based fine-granularity bounding box to merge overly fine-grained layers and filter out redundant ones. The refined dataset comprises \textbf{160K} posters, encompassing a total of \textbf{2.6M} layers (1.2M text and 1.4M image/decorative).

\subsection{Experimental Setup}
\label{sec:exp_setup}
\vspace{-1mm}

\textbf{Implementation:} Our design model employs Qwen-2.5-VL-7B-Instruct \cite{qwen2.5-VL} as backbone; the generative agent employs Qwen-Image-Edit-2509 \cite{wu2025qwenimagetechnicalreport} as T2I model; the reception model uses Qwen-2.5-7B \cite{qwen2.5}; and RLAF utilizes VisualQuality-R1 \cite{wu2025visualquality} as reward model. All experiments run on 8×RTX H20 GPUs. 
\\
\textbf{Baselines:} We compare against: (1) commercial platforms (Microsoft Designer, Nano-Banana); (2) academic SOTAs (LaDeCo \cite{lin2025elements}, CreatiPoster \cite{zhang2025creatiposter}); and (3) reasoning models (Gemini 2.5 Pro \cite{comanici2025gemini}, Qwen-VL-2.5-72B-Instruct \cite{qwen2.5-VL}).
\\
\textbf{Metrics:} Following expert consultation, we evaluate the quality of the posters generated via ratings on key metrics for graphic design: Layout Rationality \cite{feng2024layoutgpt, zheng2023layoutdiffusion}, Text Legibility \cite{chen2023textdiffuser}, Element Preservation \cite{li2023gligen}, Style Consistency \cite{sohn2023styledrop}, Instruction Following \cite{saharia2022photorealistic} and Visual Appeal \cite{saharia2022photorealistic}  for holistic poster quality evaluation, complemented by quantitative IoU, Inverse order pair ratio (IOPR) \cite{cheng2025graphic}, and Aspect Ratio Distortion (ARD) \cite{zheng2020distance} for ablation study.
\\
\textbf{Evaluation Procedure:} We performed human evaluation, supplemented by GPT-5 \cite{openai2025gpt5} as an extra reliable evaluator. For human evaluation, we conducted pairwise, binary-choice comparisons against each baseline. We sampled 25 examples per baseline, all generated from identical prompts and fully-provided assets. We collected 5 judgments per example, totaling 750 responses from over 40 evaluators with graphic design backgrounds. For GPT-5 evaluation, we used in-context learning to align the model with our scoring criteria, ensuring a strict and fair assessment of all designs. We prompted GPT-5 to evaluate all results ten times, taking the average of its ratings as the final score for each method.
\textcolor[HTML]{F52731}{\textbf{\textit{More information about Experimental Details can be found in supplementary material.}}}
\section{Results and Analysis}
\label{sec:results}
\vspace{-1mm}

\subsection{Comparison with baselines}
\begin{figure}[t]

    \centering
    \includegraphics[width=\columnwidth]{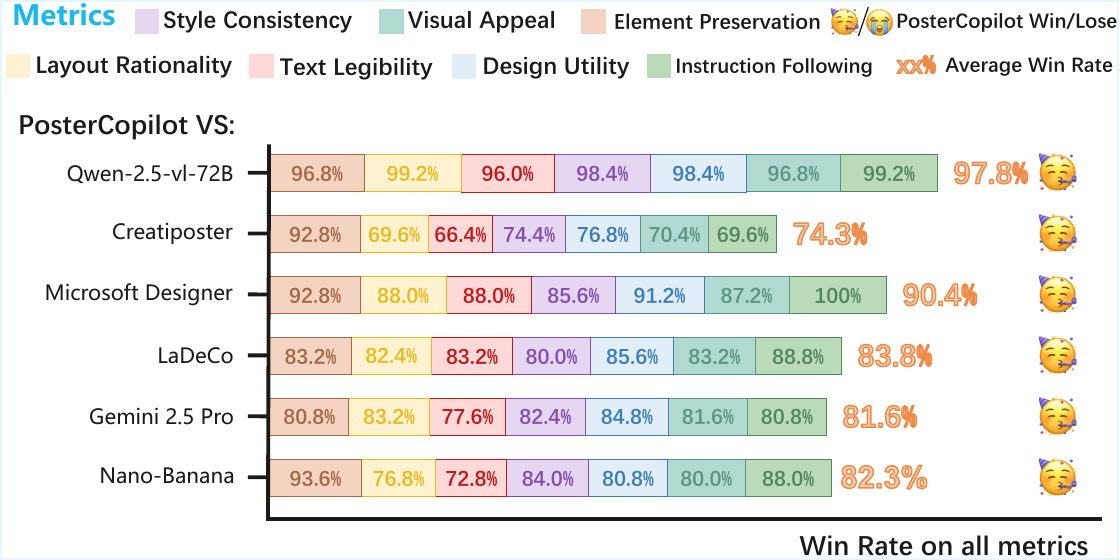}
    \captionsetup{aboveskip=0pt,belowskip=-14pt}
    \caption{Results of User-study.}
    \label{fig:user_study} 
\end{figure}

\begin{figure}[t]

    \centering
    \includegraphics[width=\columnwidth]{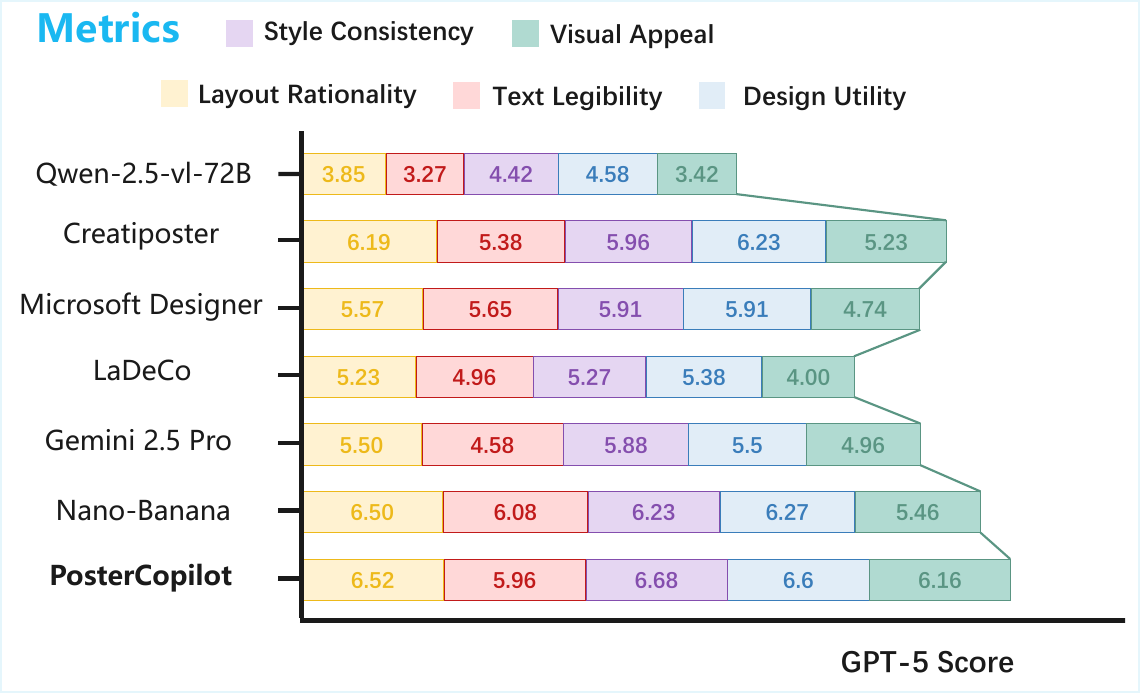}
    \captionsetup{aboveskip=0pt,belowskip=-14pt}
    \caption{Results of GPT-5 evaluation.}
    \label{fig:gpt_score} 
\end{figure}

\vspace{-1mm}
Results of human evaluation is as shown in Fig.~\ref{fig:user_study}, PosterCopilot's average win rate is well above 74\% across all baselines. While LMM-based methods such as LaDeCo perform poorly on Layout Rationality and T2I models like Nano-banana struggle with Element Preservation, PosterCopilot preserves all user-provided elements while delivering harmonious, aesthetically pleasing designs.
For GPT-5 evaluation, while GPT-5 excels at holistic quality assessment, it struggles with "instruction following" and "element preservation" as it cannot reliably process the source assets for these tasks. Consequently, these metrics were omitted from our GPT-5 evaluation. PosterCopilot's superiority in these specific areas was instead validated through our user study, which confirmed its high-fidelity performance with a dominant win rate exceeding 87\% on both. The results of GPT-5 evaluation is shown in Fig.~\ref{fig:gpt_score} . We can see that PosterCopilot decisively outperforms other methods across most metrics. PosterCopilot is slightly deficient in Text Legibility compared to Nano-Banana, because PosterCopilot prioritizes faithfully preserving all user-requested text, scaling it as needed for a harmonious layout. Nano-Banana, conversely, often achieves its legibility by simply discarding user elements—a flaw confirmed by its low Element Preservation score in our user study.

\vspace{-1mm}
\subsection{Ablation Study}
\vspace{-1mm}

The RL-VRA and RLAF phases instill professional design principles to address SFT-stage issues, including bounding box drift, element distortion, and aspect ratio errors. Evaluated using IoU, IOPR, and ARD metrics (Tab.~\ref{tab:stage_training_ablation}), RL-VRA significantly improves layout accuracy over PSFT, with further IOPR/ARD gains in RLAF. The slight IoU drop in RLAF reflects its shifted focus from ground-truth fitting to aesthetic exploration.
As detailed in Sec.~\ref{sec:rlvra}, the RL-VRA reward comprises three components: Spatial Coherence ($r_{\text{DIoU}}$), Element Fidelity ($r_{\text{size}}+r_{\text{AR}}$), and format reward. Our ablation study results in Tab.~\ref{tab:reward_ablation} on the first two rewards reveal their distinct contributions: the Spatial Coherence reward substantially enhances layout accuracy, while the Element Fidelity reward improves preservation of element sizes and proportions. Their combination yields optimal performance.
\begin{table}[t] 
\centering

\begin{subtable}{\columnwidth}
    \centering

    \resizebox{\linewidth}{!}{%
        \begin{tabular}{l ccc ccc}
        \toprule
        \multirow{2}{*}{ID} & \multicolumn{3}{c}{Training Stages} & \multirow{2}{*}{IOU$\uparrow$} & \multirow{2}{*}{IOPR$\downarrow$} & \multirow{2}{*}{ARD$\downarrow$} \\
        \cmidrule(lr){2-4}
        & PSFT        & RL-VRA     & RLAF        &         &         &         \\
        \midrule
        I     & \checkmark &            &            & 0.311     & 3.38      & 0.699     \\
        II    & \checkmark & \checkmark &            & \cellcolor[HTML]{FFCCC9}0.347     & 1.72      & 0.061     \\
        III   & \checkmark & \checkmark & \checkmark & 0.342     & \cellcolor[HTML]{FFCCC9}0.56      & \cellcolor[HTML]{FFCCC9}0.045     \\
        \bottomrule
        \end{tabular}
    }
    \caption{Ablation on training stages.}
    \label{tab:stage_training_ablation}
\end{subtable}

\vspace{0.6\baselineskip}

\begin{subtable}{\columnwidth}
    \centering
    \resizebox{\linewidth}{!}{%
        \begin{tabular}{l ccc ccc} 
        \toprule
        \multirow{2}{*}{ID} & \multicolumn{3}{c}{Layout Rewards} & \multirow{2}{*}{IOU $\uparrow$} & \multirow{2}{*}{IOPR $\downarrow$} & \multirow{2}{*}{ARD $\downarrow$} \\
        \cmidrule(lr){2-4} 
        & $r_{\text{format}}$ & $r_{\text{DIOU}}$ & $r_{\text{AR}}+r_{\text{size}}$  & & & \\ 
        \midrule
        I     & \checkmark &            &            & 0.317 & 3.29  & 0.707 \\
        II    & \checkmark & \checkmark &            & 0.339 & 1.95  & 0.734 \\
        III   & \checkmark & \checkmark & \checkmark & \cellcolor[HTML]{FFCCC9}0.347 & \cellcolor[HTML]{FFCCC9}1.72  & \cellcolor[HTML]{FFCCC9}0.061 \\
        \bottomrule
        \end{tabular}
    }
    \caption{Ablation on reward components of RL-VRA.}
    \label{tab:reward_ablation}
\end{subtable}

\captionsetup{aboveskip=3pt,belowskip=-24pt} 
\caption{Comprehensive ablation studies for training stages and reward components. We highlight the best results in red.}
\label{tab:main_ablation}
\end{table}

\vspace{-1mm}
\section{Conclusion}
\label{sec:conclusion}
\vspace{-1mm}

PosterCopilot revolutionizes automated poster design by decoupling creation into layout design and multi-round editing. Our progressive training paradigm forges the design model with geometric precision and human-like aesthetics, while a generative agent enables multi-round, layer-wise editing mirroring professional workflows. Limitations include the lack of a poster-specific aesthetic reward model and the use of standard blend modes, pointing to future work.
{
    \small
    \bibliographystyle{ieeenat_fullname}
    \bibliography{main}
}
\clearpage
\setcounter{page}{1}
\maketitlesupplementary

\section{Related Work}
\label{sec:rationale}
\subsection{Intelligent Graphic Design System}
\textbf{Single-layer Graphic Design Generation} initially relied on rule-based methods and human aesthetic constraints \cite{o2014learning, zheng2019content, yang2016automatic}, or framed the task as a constrained optimization problem \cite{lee2020neural,chen2024iris}. The paradigm shifted with the advent of text-to-image (T2I) models, driving research into enhancing the compositional capabilities of diffusion models by integrating layout information. Examples include GLIGEN \cite{li2023gligen}, LayerDiff \cite{huang2024layerdiff}, and MS-Diffusion \cite{wang2024ms}, with LayoutDiffusion \cite{zheng2023layoutdiffusion} specifically using layout as a conditioning modality. CreatiDesign \cite{zhang2025creatidesign} integrates user assets but requires pre-defined layouts. However, this single-layer approach inherently limits iterative refinement and editability, often leading to visual inconsistency and distortion in unmodified regions \cite{inoue2023layoutdm}, which increases user burden and limits usability.
\\
\textbf{Multi-layers Graphic Layout Planning} has gained attention due to its focus on real-world practicality, operating by first inferring a layout and then assembling multiple layers to offer high flexibility and editability. Early Transformer-based methods, including LayoutTransformer \cite{gupta2021layouttransformer}, BLT \cite{kong2022blt}, and LayoutDETR \cite{yu2024layoutdetr}, reframed generation as a layout prediction task, but their flat, sequential representations lacked the necessary hierarchical structure for complex designs. The subsequent rise of Vision-Language Models (VLMs) led to VLM-assisted approaches like LayoutPrompter \cite{lin2023layoutprompter} and LayoutUWNA \cite{tang2023layoutnuwa}, which use in-context learning for layout inference. PosterLLAVA \cite{yang2024posterllava} guides generation through Vision Supervised Fine-Tuning on layered designs. Other methods focus on asset integration (Graphist \cite{cheng2025graphic}), typography (POSTA \cite{chen2025posta}), or external generative capabilities (CreatiPoster \cite{zhang2025creatiposter}, COLE \cite{jia2023cole}). Crucially, these models primarily mimic static datasets rather than learning from the aesthetic quality of their own outputs. Our strategy moves beyond simple mimicry, internalizing fundamental principles of layout generation and visual aesthetics from direct generative feedback.

\subsection{Reinforcement Learning for Visually Grounded Layout Generation}
Reinforcement Learning (RL) has significantly advanced the alignment of Vision-Language Models (VLMs) with human preferences \cite{zhan2025vision,nguyen2025aligning} and enhanced their reasoning capabilities \cite{chen2025perception,ji2025enhancing}. Various VLM-based visual reward models, such as HPSv2 \cite{wu2023human}, ImageReward \cite{xu2023imagereward}, and VisualQuality-R1 \cite{wu2025visualquality}, are trained on human preference datasets to provide aesthetic feedback. However, the feedback signals from these models are often overly general and holistic, lacking specific assessments of crucial graphic design elements like layout and alignment. While AesthetiQ \cite{patnaik2025aesthetiq} utilizes Direct-Preference-Optimization (DPO) \cite{rafailov2023direct} to embed preferences in layout models, its basic feedback mechanism struggles with complex, nuanced preference signals. Our approach addresses these limitations by proposing a multi-stage Reinforcement Learning framework that directly integrates fine-grained layout principles and human aesthetic feedback into the model's learning process.

\section{Implementation details for three-stage training process}
\label{training_implementations}
Training is conducted in three stages: (1) initial PSFT phase, training for 3 epochs on 160K high-quality samples from our PosterCopilot datasets; (2) RL-VRA phase on 20K samples exhibiting complex layout rules; and (3) the final RLAF phase on 1k expert-validated samples.
\subsection{Implementation details for PSFT phase}
As shown in Eq.5 in the main text, we perturb the bounding box values of each element in the ground truth layout to conduct our PSFT training phase. This process transforms them from single, precise values into a Gaussian-like distribution, using the original value as the mean and a small parameter as the variance. Subsequently, we sample $n$ values from this distribution and then calculate the PSFT loss.

\begin{figure*}[t!]
    \centering 
    
    \begin{subfigure}[b]{0.32\textwidth} 
        \centering
        \includegraphics[width=\linewidth]{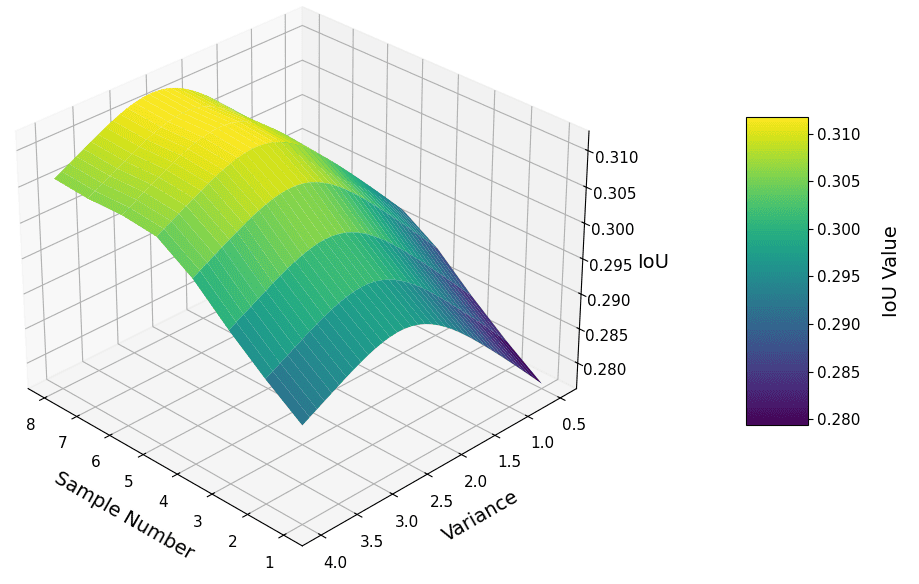} 
    \end{subfigure}
    \hfill 
    \begin{subfigure}[b]{0.32\textwidth}
        \centering
        \includegraphics[width=\linewidth]{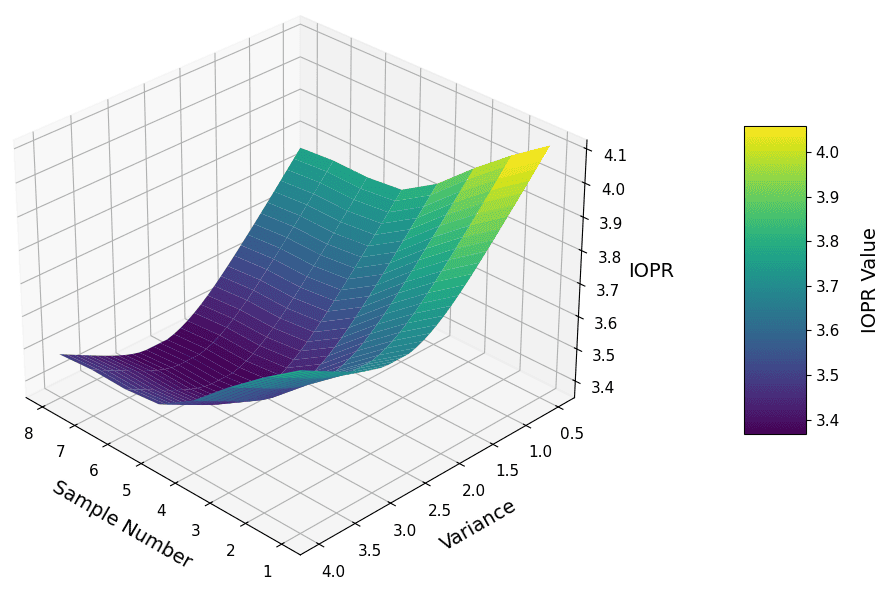}
    \end{subfigure}
    \hfill 
    \begin{subfigure}[b]{0.32\textwidth}
        \centering
        \includegraphics[width=\linewidth]{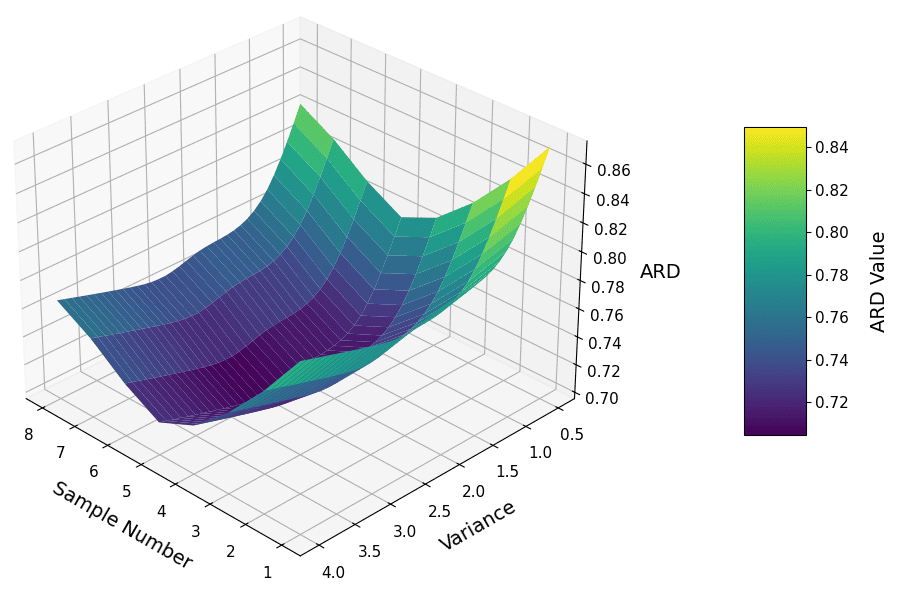}
    \end{subfigure}
    
    \captionsetup{aboveskip=2pt,belowskip=0pt}
    
    \caption{Visualization of the hyperparameter analysis for the PSFT phase.}
    \label{fig:para_psft}
\end{figure*}

Prior to the PSFT training, we conducted a grid analysis on the hyperparameters: \textbf{(1) the standard deviation $\sigma$ of the applied perturbation,} and \textbf{(2) the PSFT sampling number $n$.} We evaluated a wide range of $\sigma$ and $n$ combinations. The quality of layouts generated by design models trained with these different parameter combinations was measured using IoU, ARD, and IOPR. The resulting impact of these parameters on the PSFT stage is illustrated in Fig.~\ref{fig:para_psft}. It is evident that when the standard deviation $\sigma$ of the added perturbation is below 3.0, the model's overall performance in the PSFT phase improves as the perturbation magnitude increases. This is because adding perturbation effectively mitigates the numerical gap caused by text tokens performing regression tasks. Learning a distribution (rather than a single point) allows the design model to better grasp key layout patterns.  When the standard deviation $\sigma$ exceeds 3.0, the model's performance shows a slight degradation as $\sigma$ increases (for a fixed sampling number $n$). This is because the spread of the perturbed distribution becomes excessively large, which interferes with the model's learning of the ground truth layout.Conversely, increasing the sampling number $n$ consistently improves the design model's performance, regardless of the $\sigma$ value. However, this improvement becomes marginal once $n$ exceeds 5, and a larger $n$ also incurs a significant computational burden. Ultimately, to strike a balance between model performance and computational efficiency, we adopt $\sigma=2.5$ and $n=5$ as our final parameters.

\subsection{Implementation details for RL-VRA }
\subsubsection{Implementation details for reward design }
We use verl \cite{sheng2024hybridflow} for our reinforcement learning training phase. In RL-VRA phase we design a verifiable geometric reward as shown in Eq.10 in the main text:
\begin{equation}
\label{eq:reward_second}
r(\boldsymbol{G}) = \underbrace{r_{\text{DIoU}}}_{\text{Spatial Coherence}} + \underbrace{\lambda_{\text{size}} r_{\text{size}} + \lambda_{\text{AR}} r_{\text{AR}}}_{\text{Element Fidelity}} + r_{\text{format}}
\end{equation}
where $\lambda_{\text{size}}, \lambda_{\text{AR}} > 0$. In practice, we empirically set the weights $\lambda_{\text{size}}=0.6$ and $\lambda_{\text{AR}}=0.4$. In future work, a more fine-grained method for automatically determining individual reward weights based on the training stage may further improve the training effectiveness of RL-VRA. The specific calculations of several rewards in RL-VRA during the actual training process are as follows:

For $r_{\text{DIoU}}$, the raw DIoU metric is calculated for each element in each data sample, with a native value range of [-1.0, 1.0]. These values are then averaged to get $Mean-DIoU$. This average is transformed using the formula: 

\begin{equation}
\label{eq:transform_rlvra_diou}
r_{\text{DIoU}} = (Mean-DIoU + 1) / 2) \times 10.
\end{equation}

This mapping scales the original [-1.0, 1.0] range directly to the [0, 10] reward range, where a value of -1.0 (worst) corresponds to 0 points and +1.0 (perfect) corresponds to 10 points.

For $r_{\text{AR}}$, it's calculated from a normalized penalty. The function first computes the absolute log-difference between the predicted and ground truth aspect ratios for each layer, capping this penalty value at 1.0 (defined as cap in the following illustration). It then calculates the average negative penalty as shown in Eq.7 in the main text to get $r_{\text{AR}}^{original}$, which lies in the range [-1.0, 0]. This penalty is converted into the score using the formula: 
\begin{equation}
\label{eq:transform_rlvra_ar}
r_{\text{AR}} = ((r_{\text{AR}}^{original} + \text{cap}) / \text{cap}) \times 10
\end{equation}

This inverts the $r_{\text{AR}}^{original}$, mapping the worst-case penalty (-1.0) to a score of 0.0 and the no-penalty case (0.0) to a full score of 10.0.

The computation of the size accuracy reward ($r_{\text{size}}$) parallels the methodology used for the aspect ratio reward. First, the size inaccuracy for each layer in each data sample is quantified as shown in Eq.8 in the main text. This resulting penalty is capped at a maximum value of 1.0 (denoted as $\text{cap}$). The average of these individual penalties is then calculated across all layers, and its negative is taken, yielding $r_{\text{size\_smooth}}$. This ensures $r_{\text{size\_smooth}}$ is bounded within the range $[-1.0, 0]$, where -1.0 represents the maximum penalty. Finally, $r_{\text{size\_smooth}}$ is linearly transformed from its penalty-based range to the final 0--10 reward scale. This transformation is expressed in the following equation:

\begin{equation}
\label{eq:size_reward} %
r_{\text{size}} = \frac{(r_{\text{size\_smooth}} + \text{cap})}{\text{cap}} \times 10
\end{equation}

The format reward $r_{\text{format}}$ is a binary score designed to ensure the prediction layout $\hat{\boldsymbol{G}}$ is a valid JSON. It receives a full score of 10.0 if $\hat{\boldsymbol{G}}$ can be successfully parsed as a JSON object. If the string is malformed and results in a $\text{JSONDecodeError}$ or other parsing failure, the function immediately returns 0.0, effectively penalizing any syntactically incorrect outputs.

In summary, we have obtained a reward function that is dense, provides multi-dimensional geometric feedback, and has a maximum score of 30. This balanced reward structure is designed to provide effective visual feedback while simultaneously mitigating reward hacking. Furthermore, it prevents any single component from dominating the optimization process, which would otherwise lead to the neglect of other crucial objectives.

\begin{table}[h]
\centering
\begin{tabular}{ll}
\hline
\textbf{Hyperparameter} & \textbf{Value} \\
\hline
Learning Rate & $1 \times 10^{-6}$ \\
KL Loss Coefficient & 0.01 \\
Clip Ratio & 0.2 \\
Actor Entropy Coefficient & 0.01 \\
Training Batch Size & 96 \\
GRPO Group Size & 8 \\
Total Epochs & 1 \\
Learning Rate Optimizer & Adam \\
\hline
\end{tabular}

\caption{GRPO hyperparameter settings for RL-VRA}
\label{tab:grpo}
\end{table}

\subsubsection{GRPO Hyperparameter Settings for RL-VRA}
As shown in Tab.~\ref{tab:grpo}.

\subsection{Implementation details for RLAF}
\subsubsection{Implementation details for reward design }
$r_{\text{RLAF}}(G)$ is defined as Eq.11 in the main text:

\begin{equation}
\label{eq:transform_rlaf_reward}
r_{\text{RLAF}}(\boldsymbol{G}) =r_{\text{format}} + \lambda_{\text{aes}} r_{\text{aes}}(\boldsymbol{G})
\end{equation}
where $\lambda_{\text{aes}} > 0$. We adopt the same calculation method for $r_{\text{format}}$ as in the RL-VRA stage. We employ VisualQuality-R1 \cite{wu2025visualquality}, an evaluation model meticulously trained to align with human aesthetic preferences, as the judge model for the RLAF stage. Similarly, we modulate the contribution of $r_{\text{aes}}(\boldsymbol{G})$ via the hyperparameter $\lambda_{\text{aes}}$ to ensure a balanced configuration of reward scores. In our experiments, we set $\lambda_{\text{aes}}=2$. 

\subsubsection{GRPO hyperparameter settings for RLAF}

\begin{table}[h]
\centering
\label{tab:grpo_2}
\begin{tabular}{ll}
\hline
\textbf{Hyperparameter} & \textbf{Value} \\
\hline
Learning Rate & $5 \times 10^{-7}$ \\
KL Loss Coefficient & 0.01 \\
Clip Ratio & 0.4 \\
Actor Entropy Coefficient & 0.01 \\
Training Batch Size & 64 \\
GRPO Group Size & 4 \\
Total Epochs & 1 \\
Learning Rate Optimizer & Adam \\
\hline
\end{tabular}
\caption{GRPO hyperparameter settings for RLAF}
\end{table}

\section{More Details For Evaluation Metrics}
\subsection{Aesthetic Evaluation Metrics}
It has become a prevailing consensus among researchers in the field of graphic wdesign that relying on traditional AIGC metrics to gauge design quality is fundamentally unreasonable. While metrics like Fréchet Inception Distance (FID) and Structural Similarity Index (SSIM) are highly effective in natural image synthesis tasks, they prove inadequate when assessing the quality of graphic design \cite{zhang2025creatiposter,inoue2023layoutdm,jayasumana2024rethinking}. This inadequacy stems from a fundamental divergence in evaluation dimensions: FID and other metrics focus primarily on pixel-level fidelity and the statistical similarity of feature distributions. However, the core value of poster design lies not in the pixel-level replication of training data, but in layout topology, visual hierarchy, typographic aesthetics, and the semantic interaction among multi-modal elements \cite{zou2025fragment}.

Specifically, a vast majority of existing literature in the domain has critically argued that traditional AIGC metrics suffer from severe limitations. First, they lack the capability to perceive design rules. A generated poster might exhibit texture and color distributions highly consistent with the training set (yielding a favorable FID score), yet contain severe design accidents such as text occluding key image subjects, misalignment of elements, or imbalanced white space. While these errors are intolerable to human designers, they are often overlooked by evaluation systems based on convolutional features. Second, the calculation of these metrics is heavily influenced by the generative model's fit to the training data distribution. A low FID score merely indicates that the generated images are statistically similar to the training set, without measuring whether they are good in terms of visual appeal, the core of the graphic design domain. If the training data itself contains mediocre designs, traditional metrics may even reward outputs that mimic this mediocrity while penalizing high-quality designs that are innovative but deviate from the statistical mean. Consequently, directly applying traditional AIGC metrics fails to objectively evaluate the aesthetic value and layout quality of poster designs. There is an urgent need in this field to establish a novel evaluation system based on geometric constraints and human aesthetic perception.

Building upon the foundation of numerous distinguished prior works, we further consulted a diverse panel of experts—spanning from professional graphic designers to AI researchers. Through this process, we finalized a set of human evaluation metrics that are most suitable for assessing poster design. While the metrics are enumerated in Sec. 5.2 in the main text, owing to the limited space, their detailed descriptions are presented in Tab.~\ref{tab:evaluation_criteria}. These metrics cover all critical aspects of poster quality assessment, enabling a fair and comprehensive measurement of the final design quality.

\begin{table*}[t] 
    \centering 
    
    \begin{tabularx}{\textwidth}{m{3.0cm}<{\centering}|m{13.6cm}}
        \toprule

        \rowcolor{deepgrey}  
        \textbf{Criterion} & \quad \textbf{Description} \\
        \midrule 
        \rowcolor{lightgrey}
        \textbf{Layout Rationality} & Layout Rationality evaluates the global compositional coherence, rational element placement, clarity of visual hierarchy, and minimal occlusion of critical content. \\
        \rowcolor{deepgrey}
        \textbf{Text Legibility} & Text Legibility assesses the readability of the text design (determined by font choice, size, line spacing, and color) and the faithfulness of its rendering (sharp edges, no distortion, artifacts, or garbled characters). \\
        \rowcolor{lightgrey}
        \textbf{Asset Preservation} & Asset Preservation evaluates if all user-provided visual assets are fully retained and unaltered in the final result. \\
        \rowcolor{deepgrey}
        \textbf{Style Consistency} & Style Consistency assesses the coherence of stylistic treatment across all elements and the appropriateness of the overall visual style to the stated theme. \\
        \rowcolor{lightgrey}
        \textbf{Instruction Following} &Instruction following evaluates the fidelity to the textual specification, including the requested theme, style, layout, color scheme, and any required elements. \\
        \rowcolor{deepgrey}
        \textbf{Visual Appeal} & Overall Visual Appeal assesses the immediate aesthetic appeal and the ability to attract attention at first glance. \\
        \rowcolor{lightgrey}
        \textbf{Design Utility } & Design Utility assesses the suitability of the poster to be adopted as an initial design when facing the same practical brief (e.g., promoting the same product or theme. \\
        \bottomrule 
    \end{tabularx}

    \caption{Aesthetic evaluation metrics.} 
    \label{tab:evaluation_criteria} 
\end{table*}

\subsection{Layout evaluation metrics}
In the training phase, our method takes multiple layers decomposed from a complete poster as input, generates a layout in JSON format, and subsequently renders this layout into a poster image using rendering code. Following mainstream practices in prior literature, we employed three metrics—IoU, IOPR, and ARD—in the ablation study of the main text to directly measure the discrepancy between the predicted JSON layout and the ground truth layout. Here, we first provide the detailed calculation methods for these three metrics.

For IoU metric, \textbf{we clarify that all references to this metric throughout both the main text and the supplementary material denote the average IoU.} Specifically, we compute the IoU between each element in the layout generated by the design model and its corresponding element in the ground truth layout. The final IoU score for a poster sample is then derived by averaging the IoU values of all its constituent elements. The calculation of IoU is formally defined as follows:
\begin{equation}
    \label{eq:iou}
    \text{IoU}(B_{\text{pred}}, B_{\text{gt}}) = \frac{\text{Area}(B_{\text{pred}} \cap B_{\text{gt}})}{\text{Area}(B_{\text{pred}} \cup B_{\text{gt}})}
\end{equation}
where $N$ denotes the number of elements in the poster, and $B_{\text{pred}}^{(i)}$ and $B_{\text{gt}}^{(i)}$ represent the predicted and ground truth bounding boxes of the $i$-th element, respectively.

For the IOPR \cite{graphist2023hlg} metric, we evaluate the correctness of the predicted layer order, which is essential for maintaining visual hierarchy. IOPR quantifies the ratio of overlapping element pairs that violate the ground truth depth sequence. For a single sample with $n$ layers, it is calculated as:

\begin{equation}
\text{IOPR} = \frac{\sum_{i=1}^{n-1} \sum_{j=i+1}^{n} \mathbbm{1} \left(\mathcal{O}_j < \mathcal{O}_i \land \text{overlap}\left(i, j\right) \right)}{\sum_{i=0}^{n-1} \sum_{j=i+1}^{n} \mathbbm{1}} ,
\end{equation}
where $n$ is the number of layers in the hierarchical structure.
$\mathbbm{1}$ is an indicator function that returns 1 if the argument condition is true and 0 otherwise.
$\mathcal{O}$ denotes the output order or predicted order of the layers as determined by the model. $\mathcal{O}_i$ and $\mathcal{O}_j$ correspond to the predicted order positions of the $i^{th}$ and $j^{th}$ layers, respectively.
$\mathtt{overlap}(i, j)$ is a predicate function that determines whether the $i^{th}$ and $j^{th}$ layers overlap. 

For ARD metric, it's utilized to measure the aspect ratio distortion of the predicted bounding boxes relative to the ground truth. It is derived from the $v$ term of the Complete IoU (CIoU) \cite{zheng2020distance} metric, which is widely adopted in the industry:
\begin{equation}
v = \frac{4}{\pi^{2}} \left( \arctan \frac{w^{gt}}{h^{gt}} - \arctan \frac{w}{h} \right)^{2}
\end{equation}
where $w^{gt}$ and $h^{gt}$ denote the ground truth bounding box values, $w,h$ denote the predicted bounding box values, and $arctanh$ is one of the three tangent functions. In practice, we omitted the leading normalization term $\frac{4}{\pi^{2}}$ to make the metric differences more pronounced:

\begin{equation}
\text{ARD} =  \left( \arctan \frac{w^{gt}}{h^{gt}} - \arctan \frac{w}{h} \right)^{2}
\end{equation}

Although these quantitative layout evaluation metrics are less suited for assessing overall poster image quality compared to the aesthetic metrics introduced earlier, and are not directly applicable to single-layer generation or text-to-image models, they provide a more direct quantification of the discrepancy between generated layouts and the ground truth. Consequently, we employ these metrics specifically in the ablation study of the main text, rather than calculating them for all baselines.

\section{Supplementary Ablation Study}
Due to the limited space, the ablation study in the main text details only the primary training procedure and the reward component ablation results, demonstrating the necessity of each component and training phase. Here, we present additional ablation studies to directly validate the superiority of our PSFT phase over conventional SFT paradigms. Furthermore, we conduct a human evaluation to verify that RLAF guides the model to generate layouts more aligned with human aesthetics. This serves as an intuitive complement to the quantitative metrics presented in the main text. Except for the specific modules being ablated, all experimental settings for the ablation studies in both the main text and the supplementary material are identical to the training procedure described in Sec.~\ref{training_implementations}. 
\subsection{Ablation study for PSFT phase}
We evaluated the design model trained solely with PSFT against the one trained with standard SFT. The latter was trained exclusively on ground truth layouts without the introduction of perturbations or other augmentation measures. The results is as shown in Tab.~\ref{tab:sft_vs_psft}. We can see that the design model trained via PSFT significantly outperforms the standard SFT baseline across IoU, IOPR, and ARD metrics. This demonstrates that the PSFT strategy, by incorporating perturbations, effectively mitigates the numerical-semantic gap caused by treating numerical coordinates as text tokens for regression.

\begin{table}[t]
\centering

\resizebox{0.85\linewidth}{!}{%
    \begin{tabular}{l c c c}
    \toprule
    Method & IOU$\uparrow$ & IOPR$\downarrow$ & ARD$\downarrow$ \\
    \midrule
    SFT         & 0.285 & 4.12 & 0.851 \\
    PSFT (Ours) & \cellcolor[HTML]{FFCCC9}0.311 & \cellcolor[HTML]{FFCCC9}3.38 & \cellcolor[HTML]{FFCCC9}0.699 \\
    \bottomrule
    \end{tabular}
}
\caption{Quantitative comparison between standard SFT and our proposed PSFT. Best results are highlighted in red.}
\label{tab:sft_vs_psft}
\end{table}

\subsection{Ablation study for RLAF phase}
\begin{figure}[t]
    \centering
    \begin{subfigure}{0.32\linewidth}
        \centering
        \includegraphics[width=\linewidth]{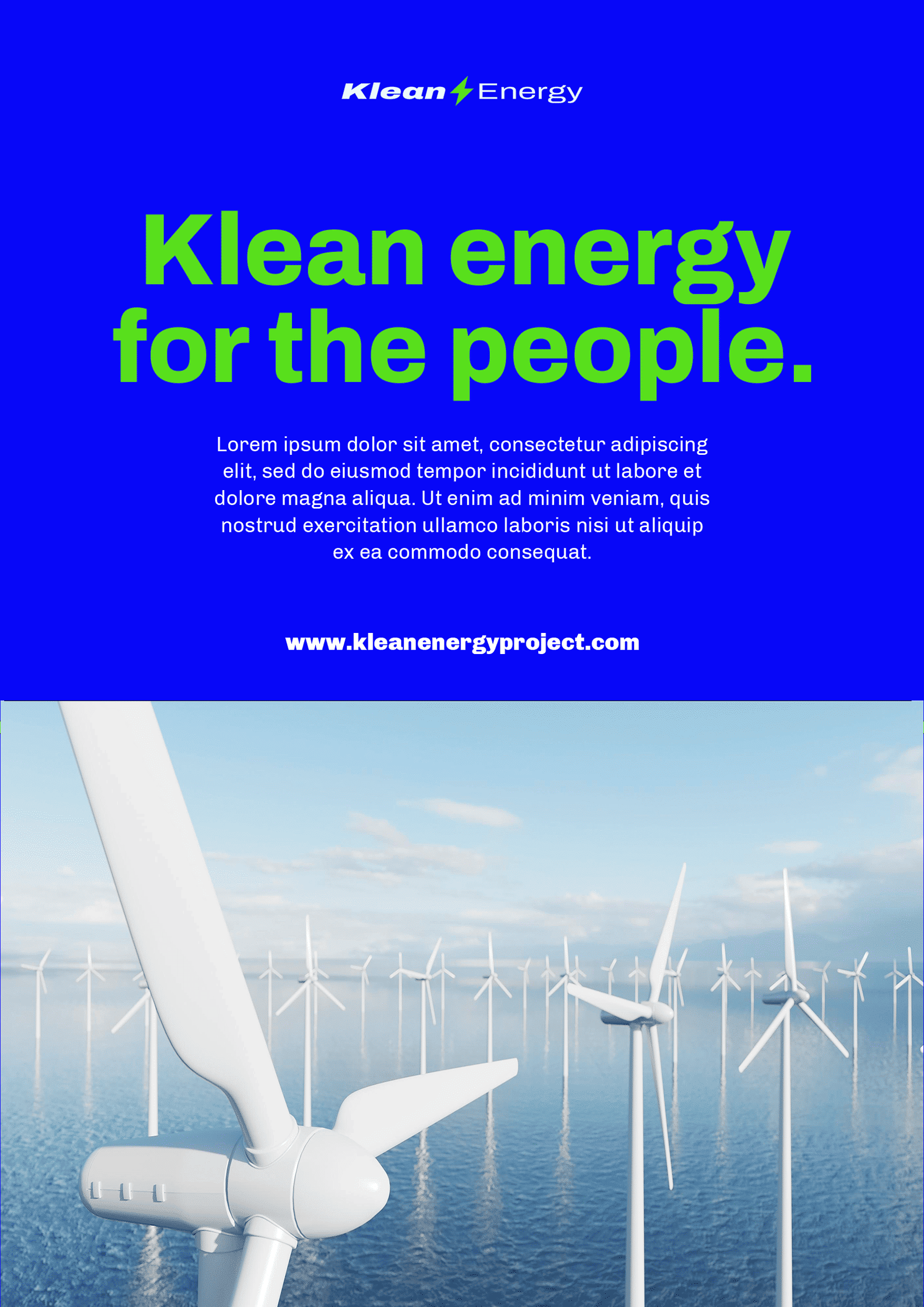}
    \end{subfigure}
    \hfill
    \begin{subfigure}{0.32\linewidth}
        \centering
        \includegraphics[width=\linewidth]{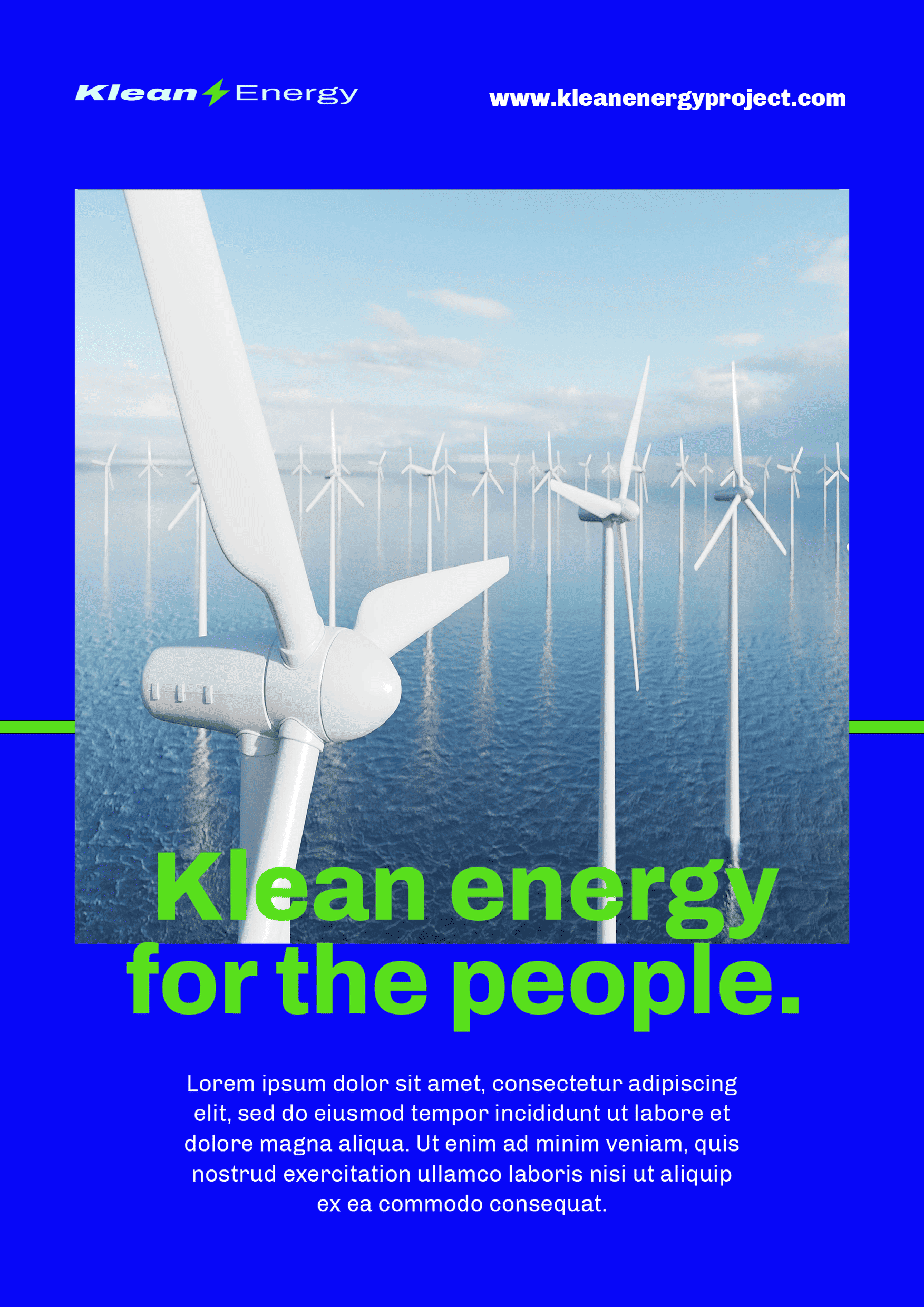}
    \end{subfigure}
    \hfill
    \begin{subfigure}{0.32\linewidth}
        \centering
        \includegraphics[width=\linewidth]{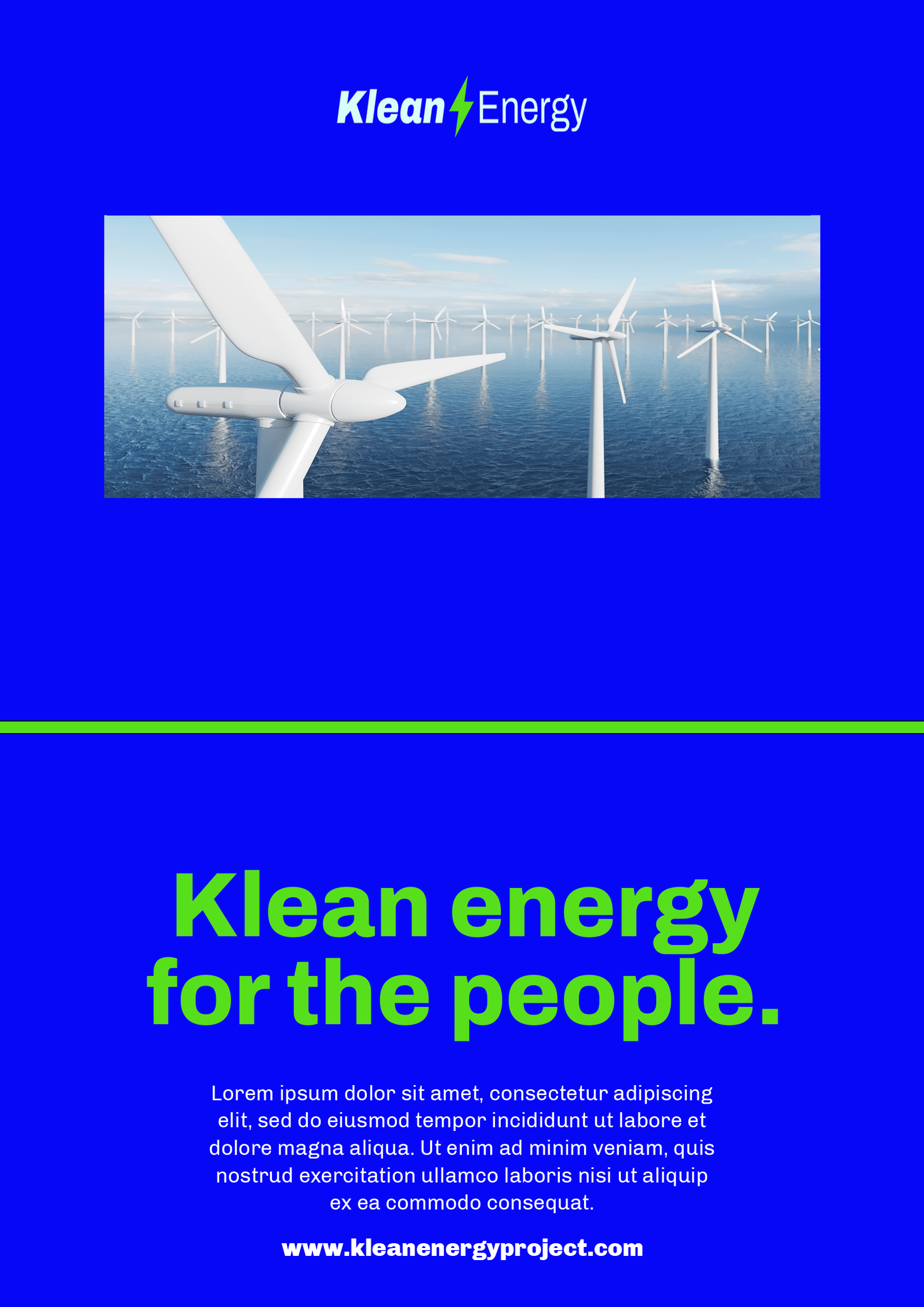}
    \end{subfigure}
    
    \vspace{-2mm} 
    
    \caption{Poster samples generated by the design model via multiple inference runs. The IoU scores against the ground truth layout are 0.87, 0.43, and 0.21, respectively. Notably, despite the varying degrees of deviation from the ground truth, all three posters align well with human aesthetics.}
    \label{fig:rlaf_three_images}
    \vspace{-3mm}
\end{figure}

\begin{figure}[ht]
    \centering
    \includegraphics[width=\columnwidth]{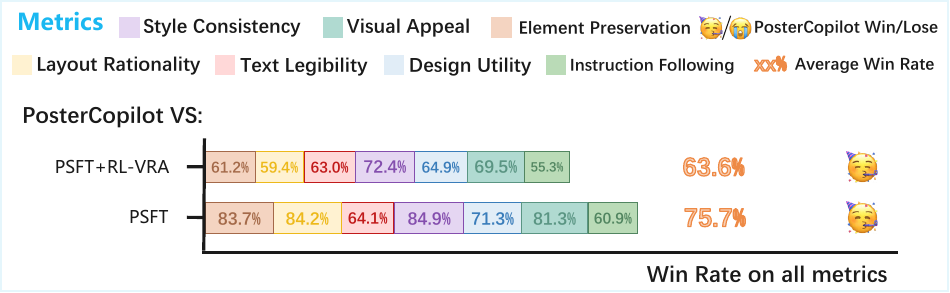}
    \captionsetup{aboveskip=0pt,belowskip=0pt}
    \caption{Human evaluation comparison of design quality metrics across different stages of our training paradigm. PosterCopilot is  trained via complete three stages. }
    \label{fig:ablation_rlaf} 
\end{figure}

Fig.~\ref{fig:rlaf_three_images} visually demonstrates the critical role of RLAF. Given that poster design is inherently driven by human aesthetics, training a design model solely to replicate ground truth layouts is insufficient. The model often generates layouts that deviate significantly from the ground truth yet remain aesthetically pleasing. In fact, layouts exhibiting greater divergence from the ground truth can sometimes yield superior aesthetic quality. We conducted an human evaluation on models trained via three progressive stages: only PSFT, PSFT + RL-VRA, and PSFT+RL-VRA+RLAF (PosterCopilot). We collected 10 inference poster samples, each of which was assessed by a panel of 15 ranging from professional graphic designers to individuals with diverse interdisciplinary backgrounds. The assessment was strictly based on the human evaluation metrics defined in the main text. Fig.~\ref{fig:ablation_rlaf} presents the evaluation results across various metrics. It is evident that the RL-VRA stage significantly enhances the layout quality and consistency of the generated designs. Building upon the previous stages, RLAF further substantially improves the visual appeal. Regarding the instruction-following capability, since the design model has already achieved a satisfactory level via training on the high-quality large-scale PosterCopilot dataset, the improvement in this metric is relatively marginal compared to other key indicators.

\section{More Details About Evaluation Procedure}
In the field of poster design, recent, open-source, and high-performing baselines capable of handling user-supplied assets are notably scarce. To ensure methodological diversity and comparison against state-of-the-art (SOTA) solutions from both academia and industry, we selected the following baselines: (1) commercial platforms (Microsoft Designer, Nano-Banana); (2) academic SOTAs (LaDeCo \cite{lin2025elements}, CreatiPoster \cite{zhang2025creatiposter}); and (3) reasoning models (Gemini 2.5 Pro \cite{comanici2025gemini}, Qwen-VL-2.5-72B-Instruct \cite{qwen2.5-VL}). As demonstrated in the main text, our comparative analysis conditions all models on identical user assets input and design prompts to generate posters. Since the baselines encompass both text-to-image (T2I) models and non-end-to-end layout generation frameworks (similar to PosterCopilot), their inference pipelines exhibit slight variations. In this section, we provide a detailed elaboration of these specific testing protocols.

\subsection{Evaluation procedure for T2I models}
Among the selected baselines, Microsoft Designer and Nano-Banana (formally known as Gemini 2.5 Flash Image) belongs to the T2I category. Notably, since its debut, Nano-Banana has garnered widespread attention within the graphic design community, distinguished by its unparalleled capabilities in multi-asset conditioned generation and multi-turn iterative editing. The evaluation procedure for T2I models is relatively straightforward. We condition the models on all provided user assets, specify the target canvas dimensions, and input the design prompt to generate the corresponding poster samples for comparison.

\subsection{Evaluation procedure for layout generation models}
The remaining methods—CreatiPoster \cite{zhang2025creatiposter}, LaDeCo \cite{lin2025elements}, Qwen-VL-2.5-72B-Instruct \cite{qwen2.5-VL}, Gemini 2.5 Pro \cite{comanici2025gemini}, and our own PosterCopilot—fall under the category of Layout Generation models. For these models, consistent with the T2I evaluation, we provide user assets and design prompts. However, we explicitly instruct the models to output the layout in JSON format. Upon obtaining the generated JSON files, we employ a unified high-precision lossless rendering script to convert the text-based layouts into final poster images for each test sample.

It's worth noting that CreatiPoster requires precise pre-defined layouts for foreground elements. To accommodate this, we provided the ground truth foreground layouts during its evaluation. Although this setup places our method at a comparative disadvantage, PosterCopilot still achieved a significant lead across all metrics in both GPT-5 evaluations and multi-dimensional human assessments. This further demonstrates PosterCopilot's robust layout reasoning capabilities while requiring minimal manual input.

\section{More Qualitative Comparisons}
\begin{figure*}[t]
    \centering
    
    \begin{subfigure}{0.95\linewidth} 
        \centering
        \includegraphics[width=\linewidth]{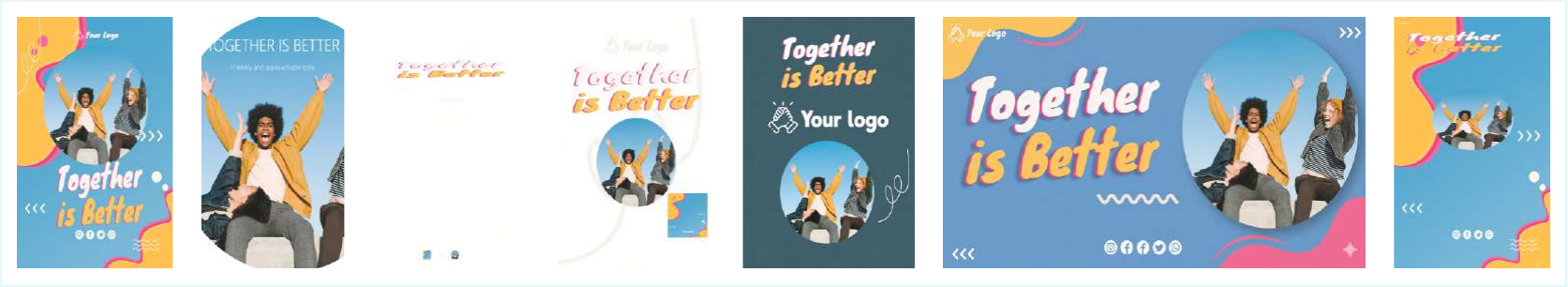}
    \end{subfigure}
    
    \vspace{0pt} 
    
    \begin{subfigure}{0.95\linewidth}
        \centering
        \includegraphics[width=\linewidth]{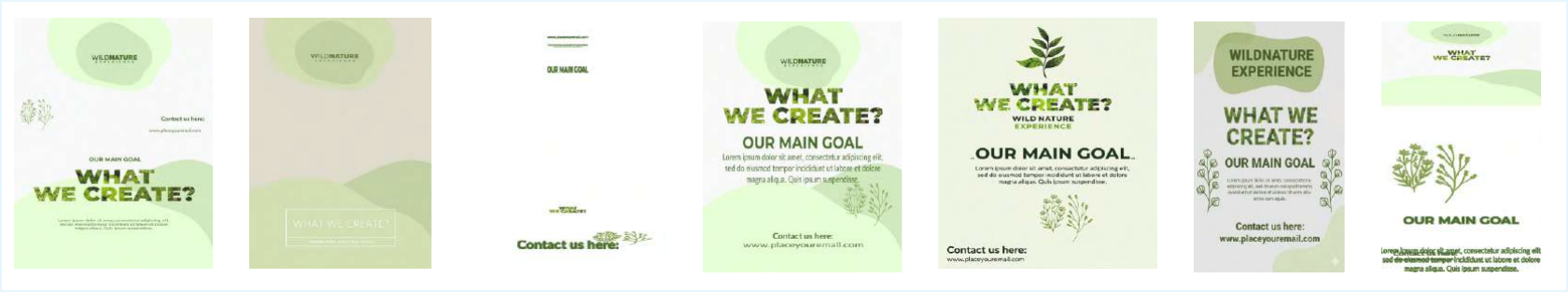}
    \end{subfigure}
    
    \vspace{0pt}
    
    \begin{subfigure}{0.95\linewidth}
        \centering
        \includegraphics[width=\linewidth]{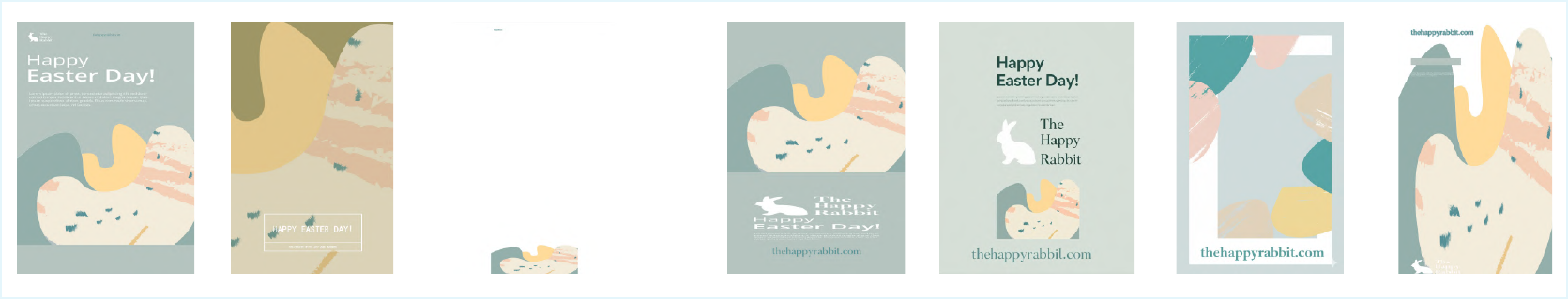}
    \end{subfigure}
    
    \vspace{0pt}

    \begin{subfigure}{0.95\linewidth}
        \centering
        \includegraphics[width=\linewidth]{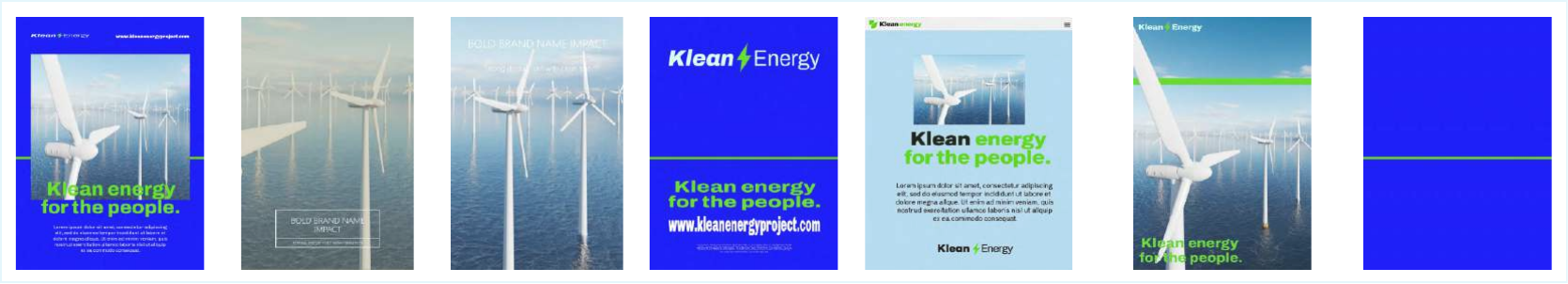}
    \end{subfigure}
    
    \vspace{0pt}

    \begin{subfigure}{0.95\linewidth}
        \centering
        \includegraphics[width=\linewidth]{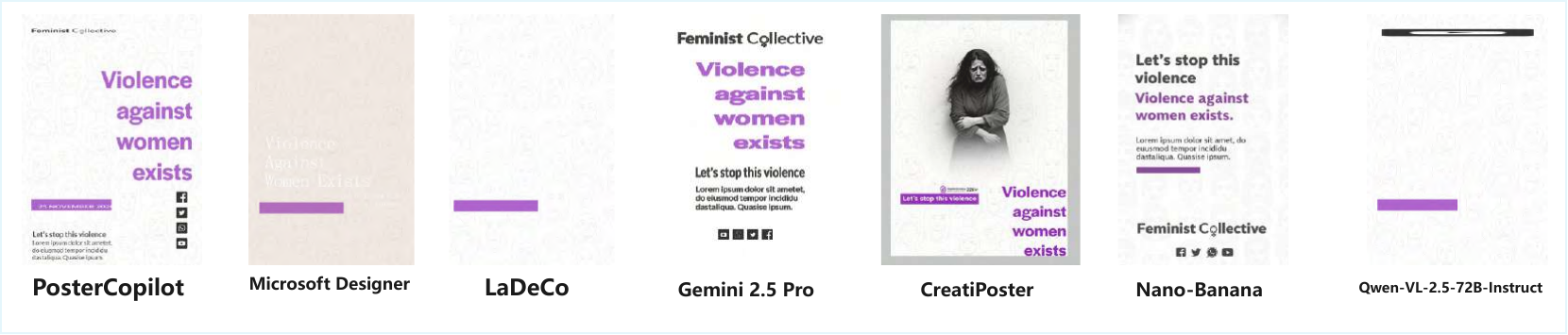}
    \end{subfigure}
    
    \caption{Visual comparison of poster composition results across all methods. Each column corresponds to a specific method, demonstrating its generation performance based on various user assets and prompts.}
    \label{fig:supp_comparison}
\end{figure*}

\begin{figure*}[t]
    \centering
    
    %
    \begin{subfigure}{0.95\linewidth} 
        \centering
        \includegraphics[width=\linewidth]{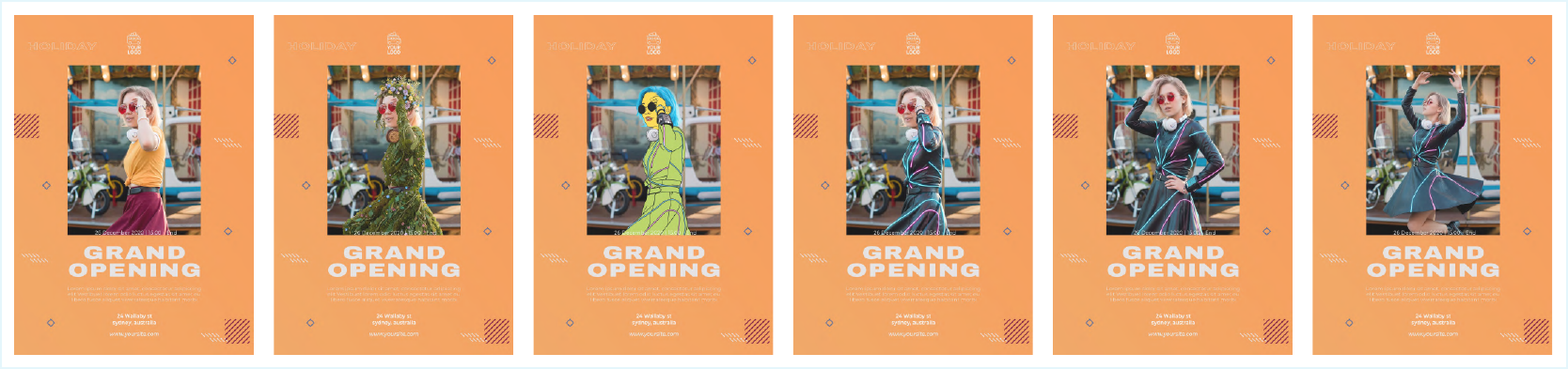}
    \end{subfigure}
    
    \vspace{0pt} %
    
    \begin{subfigure}{0.95\linewidth}
        \centering
        \includegraphics[width=\linewidth]{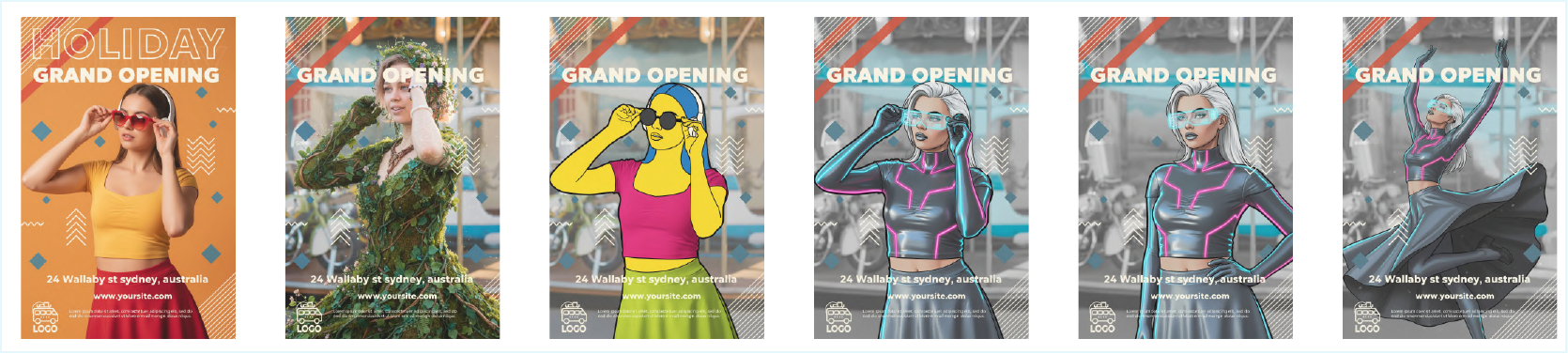}
        \caption{The first set of qualitative comparisons on single-layer editing between PosterCopilot and Nano-Banana.}
    \end{subfigure}

    \begin{subfigure}{0.95\linewidth} 
        \centering
        \includegraphics[width=\linewidth]{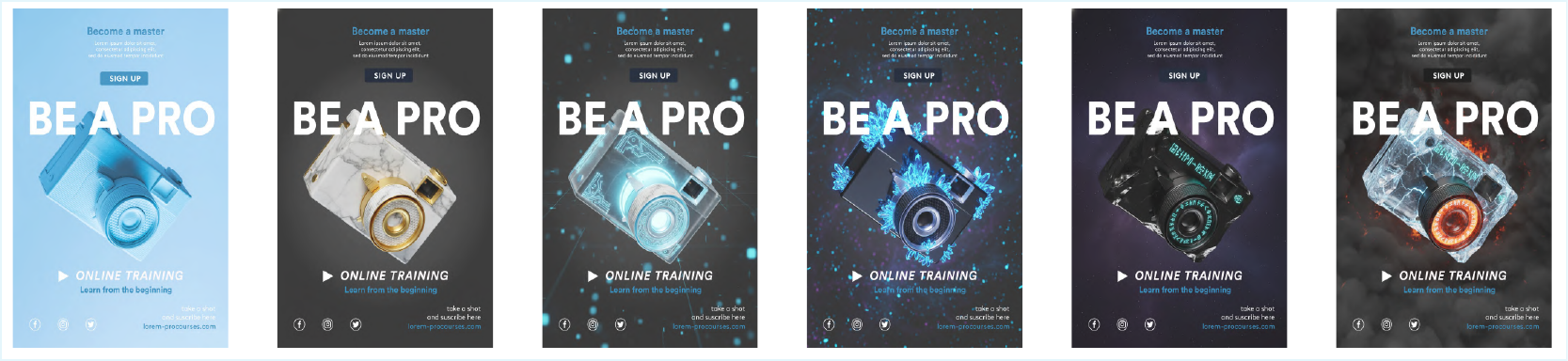}
    \end{subfigure}
    
    \vspace{0pt} %
    \begin{subfigure}{0.95\linewidth}
        \centering
        \includegraphics[width=\linewidth]{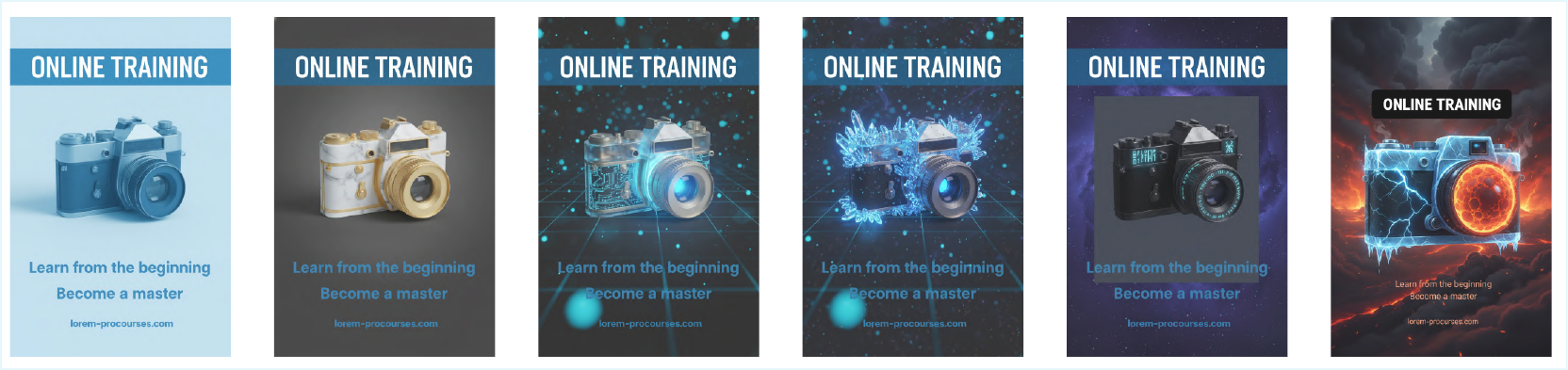}
        \caption{The second set of qualitative comparisons on single-layer editing between PosterCopilot and Nano-Banana.}
    \end{subfigure}

    \caption{Comparison of single-layer editing performance between PosterCopilot and Nano-Banana. Among all the baselines, only Nano-Banana and our PosterCopilot support the precise editing of arbitrary layers within a poster. Others either lack editing capabilities entirely or are limited to manual repositioning via dragging. Both PosterCopilot and Nano-Banana are fed with identical user assets and prompts for poster generation and multi-round edit. In each comparison, the top row shows the generation and multi-turn editing results of PosterCopilot, while the bottom row displays those of Nano-Banana. In the cases presented, the objective is to exclusively modify the background layer or the woman's appearance while leaving the rest of the poster intact. As observed, PosterCopilot faithfully preserves non-target regions throughout multi-turn editing sessions while precisely modifying the target layer. In contrast, although Nano-Banana produces impressive results initially, severe distortion occurs in other parts of the poster after just one or two refinement iterations, and unintended attributes of the subject are also altered.}
    \label{fig:supp_comparison_refine}
\end{figure*}

We provide in Fig.~\ref{fig:supp_comparison} some examples of the setting where various methods assemble posters based on complete assets. Fig.~\ref{fig:supp_comparison_refine} presents additional examples of precise single-layer editing.

\section{More details about PosterCopilot datasets}
The main text provided a key description of the PosterCopilot dataset construction pipeline. Here, we further offer more details regarding the dataset construction process and the dataset composition. 

Our datasets construction pipeline begins with the ingestion of approximately 160,000 professionally designed PSD source files collected from online stock platforms. In the initial phase, OCR Document Parsing, each PSD is exhaustively analyzed to extract all valid layers as independent PNG files. Concurrently, a JSON annotation is generated for each poster, capturing low-level metadata such as bounding boxes, stacking order, and layer type, which provides the foundation for structured supervision. 

To mitigate the fragmentation problem, the pipeline proceeds to the Parse stage, where the initial raw layers are prepared for semantic grouping. This is followed by the core Layers Merger phase. Here, the semantic cues provided by the initial OCR-based document parsing are leveraged as a data-cleaning mechanism. The merger process intelligently groups and combines excessively fine layers and concurrently discards visually insignificant ones. This crucial refinement step effectively aligns the fragmented raw layers with human visual perception, resulting in a refined annotation space focused on genuine visual elements. We present key statistics of the PosterCopilot dataset in Fig.~\ref{fig:dataset_stats}. To facilitate understanding, we also provide an example of a parsed JSON file for a representative poster instance:

\begin{figure}[t] 
    \centering
    \begin{subfigure}[b]{0.48\linewidth}
        \centering
        \includegraphics[width=\linewidth]{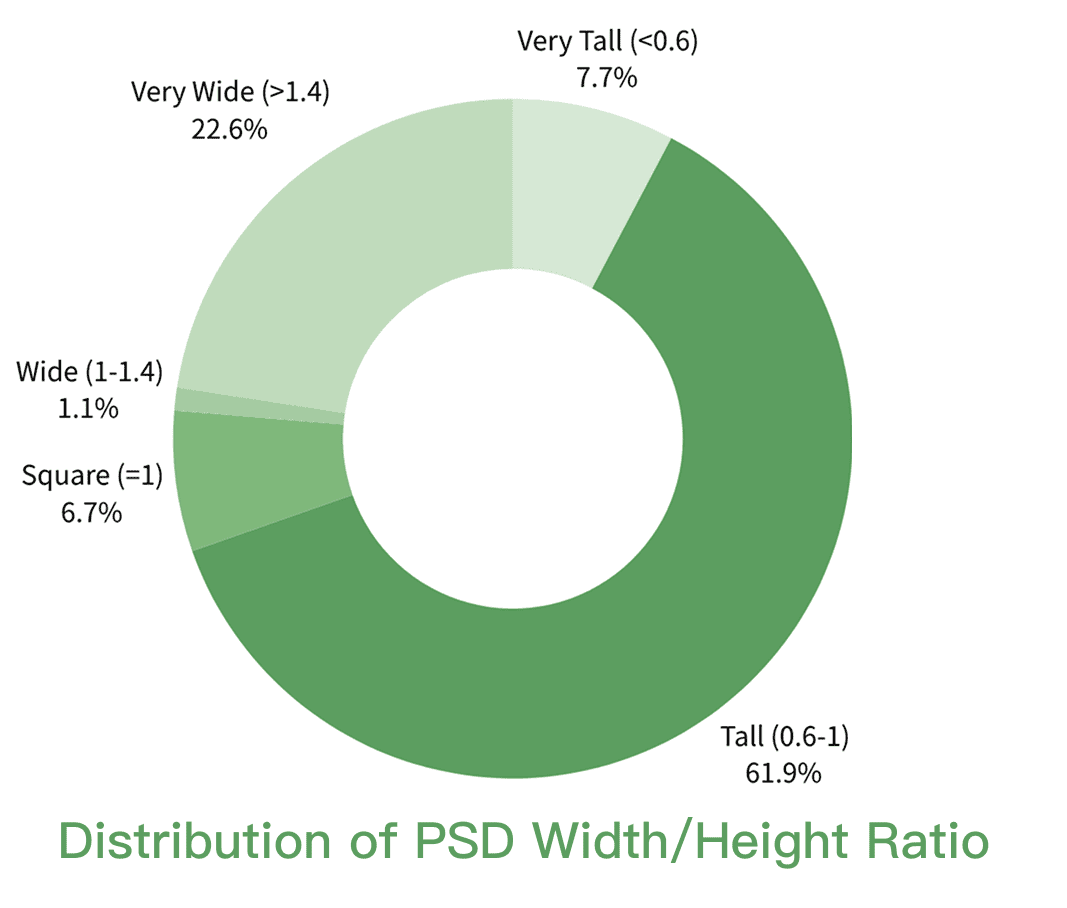}
        \label{fig:dataset_ar}
    \end{subfigure}
    \hfill 
    \begin{subfigure}[b]{0.48\linewidth}
        \centering
        \includegraphics[width=\linewidth]{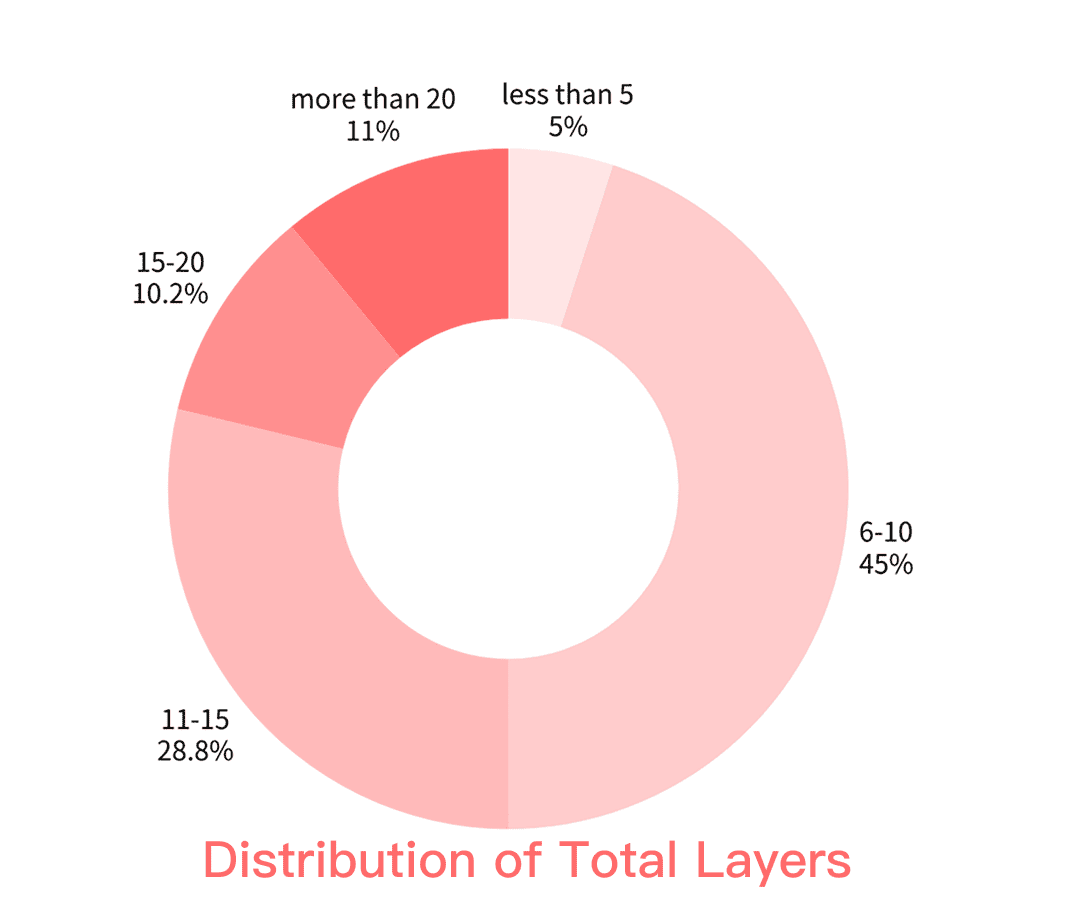}
        \label{fig:dataset_layers}
    \end{subfigure}
    
    \caption{Key statistics of the PosterCopilot dataset.} 
    \label{fig:dataset_stats}
\end{figure}

\begin{tcolorbox}[colback=gray!5, colframe=blue!60!black, title=Example of parsed JSON file,breakable]
\begin{lstlisting}[language=json, linewidth=\linewidth]
{
  "psd_file": "c:/desktop/dataset-images/freepik/freepik/Medical Poster/408589341-world-cancer-day-awareness-template/11575324.psd",
  "ocr_file": "c:/desktop/user-workspace/anonymous/psd-parsed-with-ocr/Medical Poster-408589341-11575324/ocr/11575324_ocr.json",
  "canvas_size": {
    "width": 1748,
    "height": 2480
  },
  "layers": [
    {
      "src": "World cancer day",
      "category": "type",
      "x": 144,
      "y": 537,
      "w": 1468,
      "h": 368,
      "order": 0,
      "blend_mode": "BlendMode.NORMAL",
      "opacity": 255,
      "text_info": [
        {
          "text": "WORLD CANCER DAY",
          "text_type": "PARAGRAPH",
          "font_size_px": 50.31,
          "font_family": "'Jost-ExtraBold'",
          "color_css": "rgba(96, 0, 146, 1.0)",
          "text_align": "center",
          "leading": 0.99,
          "warp": {
            "warpStyle": "b'warpNone'",
            "warpRotate": "b'Hrzn'",
            "warpValue": 0.0,
            "warpPerspective": 0.0,
            "warpPerspectiveOther": 0.0
          },
          "font-weight": "normal",
          "font-style": "normal",
          "tracking": 0.0,
          "transform": [
            4.166666666666667,
            0.0,
            0.0,
            4.166666666666667,
            -33255.49755600113,
            -32887.51407877605
          ]
        }
      ],
      "group": [
        [
          "Text",
          "World cancer day"
        ]
      ],
      "merged_layers_names": [
        "World cancer day"
      ],
      "merged_layers_num": 1,
      "merged_layers_indices": [
        0
      ],
      "is_single_layer": true,
      "files": {
        "layer": "c:/desktop/user-workspace/anonymous/psd-parsed-with-ocr/Medical Poster-408589341-11575324/merged/11575324_11_merged.png"
      },
      "ocr_info": {
        "bbox": [
          136,
          534,
          1613,
          907
        ],
        "category": "Title",
        "text": "# WORLD CANCER DAY"
      }
    },
    {
      "src": "entry free",
      "category": "type",
      "x": 74,
      "y": 119,
      "w": 156,
      "h": 117,
      "order": 1,
      "blend_mode": "BlendMode.NORMAL",
      "opacity": 255,
      "text_info": [
        {
          "text": "entry free",
          "text_type": "PARAGRAPH",
          "font_size_px": 11.4,
          "font_family": "'Montserrat-SemiBold'",
          "color_css": "rgba(96, 0, 146, 1.0)",
          "text_align": "start",
          "leading": 1.2,
          "warp": {
            "warpStyle": "b'warpNone'",
            "warpRotate": "b'Hrzn'",
            "warpValue": 0.0,
            "warpPerspective": 0.0,
            "warpPerspectiveOther": 0.0
          },
          "font-weight": "normal",
          "font-style": "normal",
          "tracking": 0.0,
          "transform": [
            4.166666666666667,
            0.0,
            0.0,
            4.166666666666667,
            -33253.49994542471,
            -32886.160441080734
          ]
        }
      ],
      "group": [
        [
          "Text",
          "entry free"
        ]
      ],
      "merged_layers_names": [
        "entry free"
      ],
      "merged_layers_num": 1,
      "merged_layers_indices": [
        1
      ],
      "is_single_layer": true,
      "files": {
        "layer": "c:/desktop/user-workspace/anonymous/psd-parsed-with-ocr/Medical Poster-408589341-11575324/merged/11575324_10_merged.png"
      },
      "ocr_info": {
        "bbox": [
          69,
          112,
          234,
          241
        ],
        "category": "Text",
        "text": "entry\nfree"
      }
    },
    {
      "src": "4/02",
      "category": "type",
      "x": 697,
      "y": 122,
      "w": 319,
      "h": 128,
      "order": 3,
      "blend_mode": "BlendMode.NORMAL",
      "opacity": 255,
      "text_info": [
        {
          "text": "4/02",
          "text_type": "PARAGRAPH",
          "font_size_px": 32.59,
          "font_family": "'Montserrat-ExtraBold'",
          "color_css": "rgba(96, 0, 146, 1.0)",
          "text_align": "start",
          "leading": 1.2,
          "warp": {
            "warpStyle": "b'warpNone'",
            "warpRotate": "b'Hrzn'",
            "warpValue": 0.0,
            "warpPerspective": 0.0,
            "warpPerspectiveOther": 0.0
          },
          "font-weight": "normal",
          "font-style": "normal",
          "tracking": 0.0,
          "transform": [
            4.166666666666667,
            0.0,
            0.0,
            4.166666666666667,
            -33255.50039401008,
            -32886.97428385417
          ]
        }
      ],
      "group": [
        [
          "Text",
          "4/02"
        ]
      ],
      "merged_layers_names": [
        "4/02"
      ],
      "merged_layers_num": 1,
      "merged_layers_indices": [
        3
      ],
      "is_single_layer": true,
      "files": {
        "layer": "c:/desktop/user-workspace/anonymous/psd-parsed-with-ocr/Medical Poster-408589341-11575324/merged/11575324_8_merged.png"
      },
      "ocr_info": {
        "bbox": [
          693,
          118,
          1021,
          251
        ],
        "category": "Text",
        "text": "4/02"
      }
    },
    
    {
      "src": "@cancer_day",
      "category": "type",
      "x": 364,
      "y": 2119,
      "w": 295,
      "h": 52,
      "order": 5,
      "blend_mode": "BlendMode.NORMAL",
      "opacity": 255,
      "text_info": [
        {
          "text": "@cancer_day",
          "text_type": "PARAGRAPH",
          "font_size_px": 12.29,
          "font_family": "'Jost-Medium'",
          "color_css": "rgba(96, 0, 146, 1.0)",
          "text_align": "start",
          "leading": 1.2,
          "warp": {
            "warpStyle": "b'warpNone'",
            "warpRotate": "b'Hrzn'",
            "warpValue": 0.0,
            "warpPerspective": 0.0,
            "warpPerspectiveOther": 0.0
          },
          "font-weight": "normal",
          "font-style": "normal",
          "tracking": 0.0,
          "transform": [
            4.166666666666667,
            0.0,
            0.0,
            4.166666666666667,
            -33254.49951986482,
            -32887.92683919271
          ]
        }
      ],
      "group": [
        [
          "Text",
          "@cancer_day"
        ]
      ],
      "merged_layers_names": [
        "@cancer_day"
      ],
      "merged_layers_num": 1,
      "merged_layers_indices": [
        5
      ],
      "is_single_layer": true,
      "files": {
        "layer": "c:/desktop/user-workspace/anonymous/psd-parsed-with-ocr/Medical Poster-408589341-11575324/merged/11575324_6_merged.png"
      },
      "ocr_info": {
        "bbox": [
          361,
          2116,
          658,
          2172
        ],
        "category": "Text",
        "text": "@cancer_day"
      }
    },
    {
      "src": "Healthy Life Avenue, 8842 Melrose st., LA,California",
      "category": "type",
      "x": 71,
      "y": 2248,
      "w": 982,
      "h": 104,
      "order": 6,
      "blend_mode": "BlendMode.NORMAL",
      "opacity": 255,
      "text_info": [
        {
          "text": "Healthy Life Avenue, 8842 Melrose st., LA,California",
          "text_type": "PARAGRAPH",
          "font_size_px": 12.29,
          "font_family": "'Jost-Medium'",
          "color_css": "rgba(96, 0, 146, 1.0)",
          "text_align": "start",
          "leading": 1.2,
          "warp": {
            "warpStyle": "b'warpNone'",
            "warpRotate": "b'Hrzn'",
            "warpValue": 0.0,
            "warpPerspective": 0.0,
            "warpPerspectiveOther": 0.0
          },
          "font-weight": "normal",
          "font-style": "normal",
          "tracking": 0.0,
          "transform": [
            4.166666666666667,
            0.0,
            0.0,
            4.166666666666667,
            -33255.49974608525,
            -32887.958521327215
          ]
        }
      ],
      "group": [
        [
          "Text",
          "Healthy Life Avenue, 8842 Melrose st., LA,California"
        ]
      ],
      "merged_layers_names": [
        "Healthy Life Avenue, 8842 Melrose st., LA,California"
      ],
      "merged_layers_num": 1,
      "merged_layers_indices": [
        6
      ],
      "is_single_layer": true,
      "files": {
        "layer": "c:/desktop/user-workspace/anonymous/psd-parsed-with-ocr/Medical Poster-408589341-11575324/merged/11575324_5_merged.png"
      },
      "ocr_info": {
        "bbox": [
          69,
          2245,
          1054,
          2357
        ],
        "category": "Text",
        "text": "Healthy Life Avenue, 8842 Melrose st., LA,-\nCalifornia"
      }
    },
    {
      "src": "www.cancerday.com",
      "category": "type",
      "x": 1222,
      "y": 2118,
      "w": 459,
      "h": 52,
      "order": 7,
      "blend_mode": "BlendMode.NORMAL",
      "opacity": 255,
      "text_info": [
        {
          "text": "www.cancerday.com",
          "text_type": "PARAGRAPH",
          "font_size_px": 12.29,
          "font_family": "'Jost-Medium'",
          "color_css": "rgba(96, 0, 146, 1.0)",
          "text_align": "right",
          "leading": 1.2,
          "warp": {
            "warpStyle": "b'warpNone'",
            "warpRotate": "b'Hrzn'",
            "warpValue": 0.0,
            "warpPerspective": 0.0,
            "warpPerspectiveOther": 0.0
          },
          "font-weight": "normal",
          "font-style": "normal",
          "tracking": 0.0,
          "transform": [
            4.166666666666667,
            0.0,
            0.0,
            4.166666666666667,
            -33254.49970463595,
            -32889.040771484375
          ]
        }
      ],
      "group": [
        [
          "Text",
          "www.cancerday.com"
        ]
      ],
      "merged_layers_names": [
        "www.cancerday.com"
      ],
      "merged_layers_num": 1,
      "merged_layers_indices": [
        7
      ],
      "is_single_layer": true,
      "files": {
        "layer": "c:/desktop/user-workspace/anonymous/psd-parsed-with-ocr/Medical Poster-408589341-11575324/merged/11575324_4_merged.png"
      },
      "ocr_info": {
        "bbox": [
          1219,
          2116,
          1680,
          2170
        ],
        "category": "Text",
        "text": "www.cancerday.com"
      }
    },
    {
      "src": "cancer knows no gender or age. get a check-up regulary.",
      "category": "type",
      "x": 366,
      "y": 362,
      "w": 1004,
      "h": 97,
      "order": 8,
      "blend_mode": "BlendMode.NORMAL",
      "opacity": 255,
      "text_info": [
        {
          "text": "CANCER KNOWS NO GENDER OR AGE. GET A CHECK-UP REGULARY.",
          "text_type": "PARAGRAPH",
          "font_size_px": 12.0,
          "font_family": "'Montserrat-BoldItalic'",
          "color_css": "rgba(96, 0, 146, 1.0)",
          "text_align": "center",
          "leading": 1.2,
          "warp": {
            "warpStyle": "b'warpNone'",
            "warpRotate": "b'Hrzn'",
            "warpValue": 0.0,
            "warpPerspective": 0.0,
            "warpPerspectiveOther": 0.0
          },
          "font-weight": "normal",
          "font-style": "normal",
          "tracking": 0.0,
          "transform": [
            4.166666666666667,
            0.0,
            0.0,
            4.166666666666667,
            -33255.501571969085,
            -32886.8654327771
          ]
        }
      ],
      "group": [
        [
          "Text",
          "cancer knows no gender or age. get a check-up regulary."
        ]
      ],
      "merged_layers_names": [
        "cancer knows no gender or age. get a check-up regulary."
      ],
      "merged_layers_num": 1,
      "merged_layers_indices": [
        8
      ],
      "is_single_layer": true,
      "files": {
        "layer": "c:/desktop/user-workspace/anonymous/psd-parsed-with-ocr/Medical Poster-408589341-11575324/merged/11575324_3_merged.png"
      },
      "ocr_info": {
        "bbox": [
          361,
          355,
          1374,
          463
        ],
        "category": "Text",
        "text": "CANCER KNOWS NO GENDER OR AGE.\nGET A CHECK-UP REGULARY."
      }
    },
    {
      "src": "Anual Scientific Cancer congress",
      "category": "type",
      "x": 373,
      "y": 993,
      "w": 920,
      "h": 116,
      "order": 9,
      "blend_mode": "BlendMode.NORMAL",
      "opacity": 255,
      "text_info": [
        {
          "text": "ANNUAL SCIENTIFIC CANCER CONGRESS",
          "text_type": "PARAGRAPH",
          "font_size_px": 14.24,
          "font_family": "'Montserrat-ExtraBold'",
          "color_css": "rgba(96, 0, 146, 1.0)",
          "text_align": "center",
          "leading": 1.2,
          "warp": {
            "warpStyle": "b'warpNone'",
            "warpRotate": "b'Hrzn'",
            "warpValue": 0.0,
            "warpPerspective": 0.0,
            "warpPerspectiveOther": 0.0
          },
          "font-weight": "normal",
          "font-style": "normal",
          "tracking": 0.0,
          "transform": [
            4.166666666666667,
            0.0,
            0.0,
            4.166666666666667,
            -33255.50023252936,
            -32886.372521938516
          ]
        }
      ],
      "group": [
        [
          "Text",
          "Anual Scientific Cancer congress"
        ]
      ],
      "merged_layers_names": [
        "Anual Scientific Cancer congress"
      ],
      "merged_layers_num": 1,
      "merged_layers_indices": [
        9
      ],
      "is_single_layer": true,
      "files": {
        "layer": "c:/desktop/user-workspace/anonymous/psd-parsed-with-ocr/Medical Poster-408589341-11575324/merged/11575324_2_merged.png"
      },
      "ocr_info": {
        "bbox": [
          370,
          986,
          1296,
          1113
        ],
        "category": "Text",
        "text": "## ANNUAL SCIENTIFIC CANCER CONGRESS"
      }
    },
    {
      "src": "Vector Smart Object",
      "category": "smartobject",
      "x": 129,
      "y": 1161,
      "w": 1355,
      "h": 889,
      "order": 10,
      "blend_mode": "BlendMode.NORMAL",
      "opacity": 255,
      "text_info": {},
      "group": [
        [
          "Design",
          "Vector Smart Object"
        ]
      ],
      "merged_layers_names": [
        "Vector Smart Object"
      ],
      "merged_layers_num": 1,
      "merged_layers_indices": [
        11
      ],
      "is_single_layer": true,
      "files": {
        "layer": "c:/desktop/user-workspace/anonymous/psd-parsed-with-ocr/Medical Poster-408589341-11575324/merged/11575324_1_merged.png"
      },
      "ocr_info": {
        "bbox": [
          127,
          1156,
          1489,
          2051
        ],
        "category": "Picture"
      }
    },
    {
      "src": "Background Layer",
      "category": "background",
      "x": 0,
      "y": 0,
      "w": 1748,
      "h": 2480,
      "order": 11,
      "blend_mode": "BlendMode.NORMAL",
      "opacity": 255,
      "text_info": {},
      "group": [
        [
          "Social Media",
          "Vector Smart Object"
        ],
        [
          "Design",
          "Vector Smart Object"
        ],
        [
          "Background",
          "Background"
        ]
      ],
      "merged_layers_names": [
        "Vector Smart Object",
        "Vector Smart Object",
        "Background"
      ],
      "merged_layers_num": 3,
      "merged_layers_indices": [
        10,
        12,
        13
      ],
      "is_single_layer": false,
      "files": {
        "layer": "c:/desktop/user-workspace/anonymous/psd-parsed-with-ocr/Medical Poster-408589341-11575324/merged/11575324_0_merged.png"
      }
    }
  ],
  "statistics": {
    "original_layers": 17,
    "valid_layers": 14,
    "merged_groups": 12,
    "excluded_layers": 0,
    "out_of_bounds_layers": 3
  }
}

\end{lstlisting}
\end{tcolorbox}


\balance
\end{document}